\documentclass{article}
\usepackage{microtype}
\usepackage{graphicx}
\usepackage{booktabs}
\usepackage{tabularx}
\usepackage{multirow}
\usepackage{amsmath}
\usepackage{enumitem}
\usepackage{amssymb}
\usepackage{xcolor}
\usepackage{mdframed}
\usepackage{listings}
\usepackage{colortbl}
\usepackage{float}
\usepackage{longtable}
\usepackage{hyperref}
\newcommand{\doi}[1]{\href{https://doi.org/#1}{doi: \nolinkurl{#1}}}

\usepackage[accepted]{mlsys2025}
\usepackage{etoolbox}
\usepackage{pifont}

\newcommand{\redcross}{\textcolor{red!70!black}{\ding{55}}}
\newcommand{\greencheck}{\textcolor{green!60!black}{\checkmark}}

\makeatletter
\patchcmd{\printAffiliationsAndNotice}
  {\par\ \\ \Notice@String}{}{}
  {\PackageError{tlaps-bench-paper}{Could not remove the proceedings notice}{}}
\makeatother

\hypersetup{pdfsubject={}}

\AddToShipoutPictureFG{%
  \AtPageLowerLeft{%
    \put(\LenToUnit{0.5\paperwidth},\LenToUnit{0.5in}){%
      \makebox(0,0){\normalfont\normalsize\color{black}\thepage}%
    }%
  }%
}

\newcommand{\bench}{TLAPS-Bench}
\newcommand{\tlaplus}{TLA$^+$}
\newcommand{\tlaps}{TLAPS}
\newcommand{\proofharness}{\texorpdfstring{\tlaplus{} Proof Harness}{TLA+ Proof Harness}}
\newcommand{\harness}{harness}
\newcommand{\tilapia}{\textsc{Tilapia}}
\newcommand{\code}[1]{\texttt{#1}}
\newcommand{\para}[1]{\smallskip\noindent\textbf{#1} }

\definecolor{discussiongreen}{rgb}{0.0,0.5,0.0}
\definecolor{resultgreen}{RGB}{40,140,70}

\makeatletter
\def\LT@makecaption#1#2#3{%
  \LT@mcol\LT@cols c{\hbox to\z@{\hss\parbox[t]\LTcapwidth{%
    \sbox\@tempboxa{{\small\sl #1{#2.}} {\small #3}}%
    \ifdim\wd\@tempboxa>\hsize
      {\small\sl #1{#2.}~}{\footnotesize #3}%
    \else
      \hbox to\hsize{\hfil\box\@tempboxa\hfil}%
    \fi
    \endgraf\vskip 4pt}\hss}}}
\makeatother

\newmdenv[
  backgroundcolor=gray!5,
  linecolor=black!60,
  innertopmargin=8pt,
  innerbottommargin=8pt,
  innerleftmargin=6pt,
  innerrightmargin=6pt,
  frametitlefont=\normalfont\bfseries\color{white},
  frametitlebackgroundcolor=black!80,
  repeatframetitle=true,
]{promptbox}
\lstdefinestyle{benchmarkprompt}{
  basicstyle=\ttfamily\normalsize,
  columns=fullflexible,
  keepspaces=true,
  breaklines=true,
  breakatwhitespace=true,
  breakindent=0pt,
  showstringspaces=false,
  upquote=true,
  literate={"}{{\char34\relax}}1
    {etcd_raft.tla}{{{\color{discussiongreen}etcd\_raft.tla}}}{13},
}

\newcommand{\numBenchSpecs}{12}
\newcommand{\numBenchTargets}{70}
\newcommand{\numFlashTargets}{15}
\newcommand{\numZabTargets}{9}
\newcommand{\numCahillTargets}{1}
\newcommand{\numTlbTargets}{2}
\newcommand{\numOpenAddressingTargets}{5}
\newcommand{\numEtcdTargets}{8}
\newcommand{\numHashicorpTargets}{6}
\newcommand{\numZooKeeperTargets}{9}
\newcommand{\numMongoTargets}{1}
\newcommand{\numCCFTargets}{2}
\newcommand{\numPirateShipTargets}{11}
\newcommand{\numWildfireTargets}{1}

\newcommand{\specFlashLOC}{623}
\newcommand{\specZabLOC}{994}
\newcommand{\specCahillLOC}{407}
\newcommand{\specTlbLOC}{361}
\newcommand{\specOpenAddressingLOC}{356}
\newcommand{\specEtcdLOC}{601}
\newcommand{\specHashicorpLOC}{656}
\newcommand{\specZooKeeperLOC}{1,840}
\newcommand{\specMongoLOC}{724}
\newcommand{\specCCFLOC}{1,144}
\newcommand{\specPirateShipLOC}{586}
\newcommand{\specWildfireLOC}{1,251}

\newcommand{\numReportedSpecIssues}{46}
\newcommand{\numFixedSpecIssues}{35}

\newcommand{\numRuntimeOswaldTargets}{3}
\newcommand{\numRuntimeWalgitTargets}{7}
\newcommand{\numRuntimeSlateDbTargets}{4}
\newcommand{\numRuntimeSailfishTargets}{3}
\newcommand{\numRuntimeBoscoTargets}{5}
\newcommand{\numRuntimeDiskQueueTargets}{1}
\newcommand{\numRuntimeDisruptorMpmcTargets}{2}
\newcommand{\numRuntimeDisruptorSpmcTargets}{2}
\newcommand{\numRuntimeBrafTargets}{14}

\newcommand{\numRuntimeGermanTargets}{7}
\newcommand{\specRuntimeBoscoLOC}{140}
\newcommand{\specRuntimeGermanLOC}{402}
\newcommand{\specRuntimeOswaldLOC}{260}
\newcommand{\specRuntimeSailfishLOC}{200}
\newcommand{\specRuntimeSlateDbLOC}{564}
\newcommand{\specRuntimeDiskQueueLOC}{1,567}
\newcommand{\specRuntimeWalgitLOC}{558}
\newcommand{\specRuntimeDisruptorMpmcLOC}{215}
\newcommand{\specRuntimeDisruptorSpmcLOC}{187}
\newcommand{\specRuntimeBrafLOC}{334}

\newcommand{\numCollectionSpecs}{95}
\newcommand{\numCollectionTargets}{265}
\newcommand{\numRetiredTargets}{195}

\newcommand{\evalFailureSyntaxTasks}{31}
\newcommand{\evalFailureIntegrityTasks}{15}
\newcommand{\evalFailureTimeoutTasks}{21}
\newcommand{\evalFailureCounterexampleTasks}{9}
\newcommand{\evalFailureUndecomposedTasks}{537}
\newcommand{\evalFailureIncompleteTasks}{1,455}

\newcommand{\evalAstraSolHighCostRatio}{3.6}

\newcommand{\passThreeDetailFlashAstraMax}{100.0\% (15/15) & 129.77 & 0.557 & 172.75 & 755.3}
\newcommand{\passThreeDetailZabAstraMax}{11.1\% (1/9) & 43.04 & 0.380 & 71.11 & 237.2}
\newcommand{\passThreeDetailCahillAstraMax}{0.0\% (0/1) & 18.60 & 0.146 & 31.53 & 77.6}
\newcommand{\passThreeDetailIvyTlbAstraMax}{50.0\% (1/2) & 43.32 & 0.238 & 60.84 & 139.7}
\newcommand{\passThreeDetailOpenAddressingAstraMax}{40.0\% (2/5) & 40.27 & 0.241 & 59.60 & 150.0}
\newcommand{\passThreeDetailEtcdAstraMax}{12.5\% (1/8) & 47.60 & 0.241 & 66.92 & 220.8}
\newcommand{\passThreeDetailHashicorpAstraMax}{16.7\% (1/6) & 17.10 & 0.135 & 28.00 & 82.0}
\newcommand{\passThreeDetailZooKeeperAstraMax}{22.2\% (2/9) & 122.50 & 0.367 & 145.98 & 229.3}
\newcommand{\passThreeDetailMongoAstraMax}{0.0\% (0/1) & 15.36 & 0.111 & 24.49 & 76.2}
\newcommand{\passThreeAstraMaxAccepted}{30}
\newcommand{\passThreeAstraMaxSuccess}{42.9}
\newcommand{\passThreeAstraMaxInput}{770.22}
\newcommand{\passThreeAstraMaxOutput}{4.270}
\newcommand{\passThreeAstraMaxCost}{1086.44}
\newcommand{\passThreeAstraMaxAgentTime}{3992.5}
\newcommand{\passThreeAstraMaxProofLOC}{13,725}
\newcommand{\passThreeAstraMaxProofLOCPerTarget}{457.5}
\newcommand{\passThreeDetailFlashAstraHigh}{100.0\% (15/15) & 114.09 & 0.426 & 146.29 & 419.5}
\newcommand{\passThreeDetailZabAstraHigh}{0.0\% (0/9) & 10.50 & 0.082 & 17.44 & 54.4}
\newcommand{\passThreeDetailCahillAstraHigh}{0.0\% (0/1) & 8.03 & 0.057 & 13.28 & 39.0}
\newcommand{\passThreeDetailIvyTlbAstraHigh}{50.0\% (1/2) & 20.14 & 0.116 & 29.84 & 93.4}
\newcommand{\passThreeDetailOpenAddressingAstraHigh}{40.0\% (2/5) & 18.52 & 0.102 & 27.16 & 61.5}
\newcommand{\passThreeDetailEtcdAstraHigh}{12.5\% (1/8) & 47.67 & 0.179 & 61.27 & 171.3}
\newcommand{\passThreeDetailHashicorpAstraHigh}{16.7\% (1/6) & 20.32 & 0.110 & 29.47 & 73.3}
\newcommand{\passThreeDetailZooKeeperAstraHigh}{0.0\% (0/9) & 17.48 & 0.097 & 25.32 & 78.9}
\newcommand{\passThreeDetailMongoAstraHigh}{0.0\% (0/1) & 6.88 & 0.051 & 11.84 & 32.2}
\newcommand{\passThreeAstraHighAccepted}{24}
\newcommand{\passThreeAstraHighSuccess}{34.3}
\newcommand{\passThreeAstraHighInput}{353.11}
\newcommand{\passThreeAstraHighOutput}{1.663}
\newcommand{\passThreeAstraHighCost}{486.57}
\newcommand{\passThreeAstraHighAgentTime}{1575.7}
\newcommand{\passThreeAstraHighProofLOC}{5,503}
\newcommand{\passThreeAstraHighProofLOCPerTarget}{229.3}
\newcommand{\passThreeDetailFlashAstraLow}{100.0\% (15/15) & 125.19 & 0.395 & 158.78 & 429.9}
\newcommand{\passThreeDetailZabAstraLow}{0.0\% (0/9) & 9.21 & 0.056 & 14.40 & 42.7}
\newcommand{\passThreeDetailCahillAstraLow}{0.0\% (0/1) & 1.90 & 0.017 & 4.18 & 13.5}
\newcommand{\passThreeDetailIvyTlbAstraLow}{50.0\% (1/2) & 25.64 & 0.125 & 35.53 & 103.6}
\newcommand{\passThreeDetailOpenAddressingAstraLow}{40.0\% (2/5) & 19.32 & 0.069 & 26.00 & 53.6}
\newcommand{\passThreeDetailEtcdAstraLow}{12.5\% (1/8) & 27.31 & 0.199 & 41.52 & 253.4}
\newcommand{\passThreeDetailHashicorpAstraLow}{16.7\% (1/6) & 18.44 & 0.095 & 26.59 & 76.4}
\newcommand{\passThreeDetailZooKeeperAstraLow}{0.0\% (0/9) & 5.43 & 0.028 & 8.92 & 21.2}
\newcommand{\passThreeDetailMongoAstraLow}{0.0\% (0/1) & 4.81 & 0.025 & 7.79 & 17.9}
\newcommand{\passThreeAstraLowAccepted}{24}
\newcommand{\passThreeAstraLowSuccess}{34.3}
\newcommand{\passThreeAstraLowInput}{272.83}
\newcommand{\passThreeAstraLowOutput}{1.162}
\newcommand{\passThreeAstraLowCost}{373.41}
\newcommand{\passThreeAstraLowAgentTime}{1262.5}
\newcommand{\passThreeAstraLowProofLOC}{7,175}
\newcommand{\passThreeAstraLowProofLOCPerTarget}{299.0}
\newcommand{\passThreeDetailFlashSolMax}{100.0\% (15/15) & 161.15 & 0.702 & 37.27 & 510.7}
\newcommand{\passThreeDetailZabSolMax}{0.0\% (0/9) & 13.05 & 0.160 & 3.66 & 77.5}
\newcommand{\passThreeDetailCahillSolMax}{0.0\% (0/1) & 36.71 & 0.296 & 8.01 & 172.4}
\newcommand{\passThreeDetailIvyTlbSolMax}{50.0\% (1/2) & 61.77 & 0.407 & 11.98 & 216.3}
\newcommand{\passThreeDetailOpenAddressingSolMax}{100.0\% (5/5) & 275.38 & 1.194 & 56.93 & 846.1}
\newcommand{\passThreeDetailEtcdSolMax}{37.5\% (3/8) & 189.90 & 0.696 & 38.60 & 479.1}
\newcommand{\passThreeDetailHashicorpSolMax}{16.7\% (1/6) & 29.26 & 0.259 & 6.74 & 157.9}
\newcommand{\passThreeDetailZooKeeperSolMax}{0.0\% (0/9) & 33.22 & 0.266 & 7.58 & 131.3}
\newcommand{\passThreeDetailMongoSolMax}{0.0\% (0/1) & 23.33 & 0.183 & 5.42 & 109.8}
\newcommand{\passThreeSolMaxAccepted}{31}
\newcommand{\passThreeSolMaxSuccess}{44.3}
\newcommand{\passThreeSolMaxInput}{1456.14}
\newcommand{\passThreeSolMaxOutput}{7.730}
\newcommand{\passThreeSolMaxCost}{328.86}
\newcommand{\passThreeSolMaxAgentTime}{5917.4}
\newcommand{\passThreeSolMaxProofLOC}{15,619}
\newcommand{\passThreeSolMaxProofLOCPerTarget}{503.8}
\newcommand{\passThreeDetailFlashSolHigh}{100.0\% (15/15) & 119.17 & 0.509 & 24.92 & 338.1}
\newcommand{\passThreeDetailZabSolHigh}{0.0\% (0/9) & 15.49 & 0.140 & 3.83 & 94.9}
\newcommand{\passThreeDetailCahillSolHigh}{0.0\% (0/1) & 19.41 & 0.136 & 4.32 & 82.3}
\newcommand{\passThreeDetailIvyTlbSolHigh}{50.0\% (1/2) & 62.02 & 0.275 & 10.25 & 175.3}
\newcommand{\passThreeDetailOpenAddressingSolHigh}{40.0\% (2/5) & 30.05 & 0.168 & 6.01 & 93.6}
\newcommand{\passThreeDetailEtcdSolHigh}{12.5\% (1/8) & 63.50 & 0.272 & 10.59 & 265.2}
\newcommand{\passThreeDetailHashicorpSolHigh}{16.7\% (1/6) & 17.87 & 0.121 & 3.81 & 66.7}
\newcommand{\passThreeDetailZooKeeperSolHigh}{0.0\% (0/9) & 23.43 & 0.205 & 5.57 & 195.6}
\newcommand{\passThreeDetailMongoSolHigh}{0.0\% (0/1) & 11.50 & 0.071 & 2.48 & 42.2}
\newcommand{\passThreeSolHighAccepted}{24}
\newcommand{\passThreeSolHighSuccess}{34.3}
\newcommand{\passThreeSolHighInput}{635.57}
\newcommand{\passThreeSolHighOutput}{3.312}
\newcommand{\passThreeSolHighCost}{134.75}
\newcommand{\passThreeSolHighAgentTime}{2725.0}
\newcommand{\passThreeSolHighProofLOC}{5,967}
\newcommand{\passThreeSolHighProofLOCPerTarget}{248.6}
\newcommand{\passThreeDetailFlashSolLow}{100.0\% (15/15) & 209.83 & 0.798 & 36.96 & 665.6}
\newcommand{\passThreeDetailZabSolLow}{0.0\% (0/9) & 14.65 & 0.089 & 3.04 & 53.3}
\newcommand{\passThreeDetailCahillSolLow}{0.0\% (0/1) & 2.40 & 0.023 & 0.81 & 14.1}
\newcommand{\passThreeDetailIvyTlbSolLow}{50.0\% (1/2) & 45.09 & 0.210 & 8.26 & 168.8}
\newcommand{\passThreeDetailOpenAddressingSolLow}{40.0\% (2/5) & 17.45 & 0.085 & 3.33 & 44.0}
\newcommand{\passThreeDetailEtcdSolLow}{12.5\% (1/8) & 57.87 & 0.286 & 9.89 & 199.5}
\newcommand{\passThreeDetailHashicorpSolLow}{16.7\% (1/6) & 48.53 & 0.195 & 8.04 & 164.6}
\newcommand{\passThreeDetailZooKeeperSolLow}{0.0\% (0/9) & 10.27 & 0.060 & 2.20 & 41.6}
\newcommand{\passThreeDetailMongoSolLow}{0.0\% (0/1) & 9.95 & 0.057 & 2.06 & 29.9}
\newcommand{\passThreeSolLowAccepted}{25}
\newcommand{\passThreeSolLowSuccess}{35.7}
\newcommand{\passThreeSolLowInput}{549.23}
\newcommand{\passThreeSolLowOutput}{2.238}
\newcommand{\passThreeSolLowCost}{97.75}
\newcommand{\passThreeSolLowAgentTime}{1795.6}
\newcommand{\passThreeSolLowProofLOC}{8,834}
\newcommand{\passThreeSolLowProofLOCPerTarget}{353.4}
\newcommand{\passThreeDetailFlashTerraMax}{73.3\% (11/15) & 2054.57 & 6.003 & 555.88 & 4890.6}
\newcommand{\passThreeDetailZabTerraMax}{0.0\% (0/9) & 2220.78 & 6.001 & 580.62 & 4415}
\newcommand{\passThreeDetailCahillTerraMax}{0.0\% (0/1) & 2548.60 & 5.996 & 644.28 & 6256.9}
\newcommand{\passThreeDetailIvyTlbTerraMax}{50.0\% (1/2) & 1922.13 & 6.002 & 515.81 & 3097.6}
\newcommand{\passThreeDetailOpenAddressingTerraMax}{40.0\% (2/5) & 2266.44 & 6.005 & 594.60 & 6356.0}
\newcommand{\passThreeDetailEtcdTerraMax}{0.0\% (0/8) & 2998.18 & 6.006 & 750.54 & 9107.1}
\newcommand{\passThreeDetailHashicorpTerraMax}{16.7\% (1/6) & 1822.03 & 4.606 & 476.72 & 4625.9}
\newcommand{\passThreeDetailZooKeeperTerraMax}{0.0\% (0/9) & 2255.79 & 6.002 & 587.39 & 5290.9}
\newcommand{\passThreeDetailMongoTerraMax}{0.0\% (0/1) & 761.25 & 2.530 & 207.59 & 2349.4}
\newcommand{\passThreeTerraMaxAccepted}{15}
\newcommand{\passThreeTerraMaxSuccess}{21.4}
\newcommand{\passThreeTerraMaxInput}{21475.87}
\newcommand{\passThreeTerraMaxOutput}{56.232}
\newcommand{\passThreeTerraMaxCost}{5602.15}
\newcommand{\passThreeTerraMaxAgentTime}{52481}
\newcommand{\passThreeTerraMaxProofLOC}{15,448}
\newcommand{\passThreeTerraMaxProofLOCPerTarget}{1,029.9}
\newcommand{\passThreeDetailFlashTerraHigh}{0.0\% (0/15) & 12.89 & 0.062 & 4.34 & 50.5}
\newcommand{\passThreeDetailZabTerraHigh}{0.0\% (0/9) & 9.53 & 0.046 & 3.20 & 35.1}
\newcommand{\passThreeDetailCahillTerraHigh}{0.0\% (0/1) & 6.93 & 0.048 & 2.57 & 40.4}
\newcommand{\passThreeDetailIvyTlbTerraHigh}{0.0\% (0/2) & 16.65 & 0.074 & 5.13 & 54.3}
\newcommand{\passThreeDetailOpenAddressingTerraHigh}{0.0\% (0/5) & 10.95 & 0.060 & 3.62 & 32.1}
\newcommand{\passThreeDetailEtcdTerraHigh}{0.0\% (0/8) & 5.44 & 0.042 & 2.09 & 28.3}
\newcommand{\passThreeDetailHashicorpTerraHigh}{0.0\% (0/6) & 9.53 & 0.046 & 3.20 & 40.6}
\newcommand{\passThreeDetailZooKeeperTerraHigh}{0.0\% (0/9) & 15.98 & 0.052 & 4.82 & 56.1}
\newcommand{\passThreeDetailMongoTerraHigh}{0.0\% (0/1) & 11.86 & 0.058 & 3.82 & 52.6}
\newcommand{\passThreeTerraHighAccepted}{0}
\newcommand{\passThreeTerraHighSuccess}{0.0}
\newcommand{\passThreeTerraHighInput}{126.85}
\newcommand{\passThreeTerraHighOutput}{0.635}
\newcommand{\passThreeTerraHighCost}{42.00}
\newcommand{\passThreeTerraHighAgentTime}{528.6}
\newcommand{\passThreeDetailFlashTerraLow}{0.0\% (0/15) & 3.96 & 0.021 & 1.49 & 21.0}
\newcommand{\passThreeDetailZabTerraLow}{0.0\% (0/9) & 2.64 & 0.018 & 1.16 & 17.9}
\newcommand{\passThreeDetailCahillTerraLow}{0.0\% (0/1) & 3.21 & 0.019 & 1.29 & 25.1}
\newcommand{\passThreeDetailIvyTlbTerraLow}{0.0\% (0/2) & 4.53 & 0.033 & 1.73 & 29.9}
\newcommand{\passThreeDetailOpenAddressingTerraLow}{0.0\% (0/5) & 2.85 & 0.020 & 1.21 & 15.3}
\newcommand{\passThreeDetailEtcdTerraLow}{0.0\% (0/8) & 2.85 & 0.020 & 1.26 & 18.2}
\newcommand{\passThreeDetailHashicorpTerraLow}{0.0\% (0/6) & 2.97 & 0.020 & 1.24 & 20.7}
\newcommand{\passThreeDetailZooKeeperTerraLow}{0.0\% (0/9) & 2.05 & 0.016 & 0.98 & 14.8}
\newcommand{\passThreeDetailMongoTerraLow}{0.0\% (0/1) & 1.93 & 0.016 & 0.90 & 10.3}
\newcommand{\passThreeTerraLowAccepted}{0}
\newcommand{\passThreeTerraLowSuccess}{0.0}
\newcommand{\passThreeTerraLowInput}{35.45}
\newcommand{\passThreeTerraLowOutput}{0.243}
\newcommand{\passThreeTerraLowCost}{14.78}
\newcommand{\passThreeTerraLowAgentTime}{215.2}
\newcommand{\passThreeDetailFlashLunaMax}{26.7\% (4/15) & 194.40 & 0.861 & 2.64 & 359.5}
\newcommand{\passThreeDetailZabLunaMax}{0.0\% (0/9) & 105.81 & 0.514 & 1.44 & 164.6}
\newcommand{\passThreeDetailCahillLunaMax}{0.0\% (0/1) & 23.30 & 0.183 & 0.37 & 65.6}
\newcommand{\passThreeDetailIvyTlbLunaMax}{50.0\% (1/2) & 995.78 & 3.649 & 13.03 & 2475.9}
\newcommand{\passThreeDetailOpenAddressingLunaMax}{40.0\% (2/5) & 702.70 & 2.682 & 9.13 & 1452.2}
\newcommand{\passThreeDetailEtcdLunaMax}{0.0\% (0/8) & 18.80 & 0.190 & 0.33 & 65.1}
\newcommand{\passThreeDetailHashicorpLunaMax}{0.0\% (0/6) & 69.40 & 0.384 & 0.98 & 111.4}
\newcommand{\passThreeDetailZooKeeperLunaMax}{0.0\% (0/9) & 23.73 & 0.151 & 0.36 & 54.1}
\newcommand{\passThreeDetailMongoLunaMax}{0.0\% (0/1) & 31.62 & 0.275 & 0.53 & 74.1}
\newcommand{\passThreeLunaMaxAccepted}{7}
\newcommand{\passThreeLunaMaxSuccess}{10.0}
\newcommand{\passThreeLunaMaxInput}{2338.15}
\newcommand{\passThreeLunaMaxOutput}{9.586}
\newcommand{\passThreeLunaMaxCost}{31.11}
\newcommand{\passThreeLunaMaxAgentTime}{5285.4}
\newcommand{\passThreeLunaMaxProofLOC}{11,497}
\newcommand{\passThreeLunaMaxProofLOCPerTarget}{1,642.4}
\newcommand{\passThreeDetailFlashLunaHigh}{0.0\% (0/15) & 5.43 & 0.036 & 0.09 & 17.9}
\newcommand{\passThreeDetailZabLunaHigh}{0.0\% (0/9) & 4.19 & 0.023 & 0.07 & 31.9}
\newcommand{\passThreeDetailCahillLunaHigh}{0.0\% (0/1) & 1.57 & 0.018 & 0.04 & 8.3}
\newcommand{\passThreeDetailIvyTlbLunaHigh}{0.0\% (0/2) & 1.86 & 0.019 & 0.04 & 18.3}
\newcommand{\passThreeDetailOpenAddressingLunaHigh}{0.0\% (0/5) & 4.96 & 0.040 & 0.10 & 37.6}
\newcommand{\passThreeDetailEtcdLunaHigh}{0.0\% (0/8) & 1.52 & 0.019 & 0.04 & 7.0}
\newcommand{\passThreeDetailHashicorpLunaHigh}{0.0\% (0/6) & 1.66 & 0.019 & 0.04 & 14.8}
\newcommand{\passThreeDetailZooKeeperLunaHigh}{0.0\% (0/9) & 3.25 & 0.019 & 0.06 & 12.4}
\newcommand{\passThreeDetailMongoLunaHigh}{0.0\% (0/1) & 2.27 & 0.021 & 0.05 & 10.0}
\newcommand{\passThreeLunaHighAccepted}{0}
\newcommand{\passThreeLunaHighSuccess}{0.0}
\newcommand{\passThreeLunaHighInput}{33.98}
\newcommand{\passThreeLunaHighOutput}{0.284}
\newcommand{\passThreeLunaHighCost}{0.69}
\newcommand{\passThreeLunaHighAgentTime}{200.3}
\newcommand{\passThreeDetailFlashLunaLow}{0.0\% (0/15) & 1.14 & 0.005 & 0.03 & 2.8}
\newcommand{\passThreeDetailZabLunaLow}{0.0\% (0/9) & 1.44 & 0.005 & 0.03 & 4.1}
\newcommand{\passThreeDetailCahillLunaLow}{0.0\% (0/1) & 0.61 & 0.003 & 0.02 & 3.0}
\newcommand{\passThreeDetailIvyTlbLunaLow}{0.0\% (0/2) & 1.69 & 0.008 & 0.04 & 4.1}
\newcommand{\passThreeDetailOpenAddressingLunaLow}{0.0\% (0/5) & 0.55 & 0.005 & 0.01 & 1.4}
\newcommand{\passThreeDetailEtcdLunaLow}{0.0\% (0/8) & 0.75 & 0.005 & 0.02 & 2.0}
\newcommand{\passThreeDetailHashicorpLunaLow}{0.0\% (0/6) & 0.76 & 0.004 & 0.02 & 1.8}
\newcommand{\passThreeDetailZooKeeperLunaLow}{0.0\% (0/9) & 1.06 & 0.005 & 0.02 & 2.6}
\newcommand{\passThreeDetailMongoLunaLow}{0.0\% (0/1) & 0.70 & 0.006 & 0.02 & 1.9}
\newcommand{\passThreeLunaLowAccepted}{0}
\newcommand{\passThreeLunaLowSuccess}{0.0}
\newcommand{\passThreeLunaLowInput}{13.81}
\newcommand{\passThreeLunaLowOutput}{0.070}
\newcommand{\passThreeLunaLowCost}{0.30}
\newcommand{\passThreeLunaLowAgentTime}{45.6}
\newcommand{\passThreeSolCahillLowRequests}{72}
\newcommand{\passThreeSolCahillLowMeanReasoning}{207}
\newcommand{\passThreeSolCahillLowCost}{0.81}
\newcommand{\passThreeSolCahillHighRequests}{196}
\newcommand{\passThreeSolCahillHighMeanReasoning}{442}
\newcommand{\passThreeSolCahillHighCost}{4.32}
\newcommand{\passThreeSolCahillMaxRequests}{277}
\newcommand{\passThreeSolCahillMaxMeanReasoning}{727}
\newcommand{\passThreeSolCahillMaxCost}{8.01}
\newcommand{\passThreeSolFlashLowRequests}{1,400}
\newcommand{\passThreeSolFlashLowMeanReasoning}{389}
\newcommand{\passThreeSolFlashLowCost}{36.96}
\newcommand{\passThreeSolFlashHighRequests}{691}
\newcommand{\passThreeSolFlashHighMeanReasoning}{499}
\newcommand{\passThreeSolFlashHighCost}{24.92}
\newcommand{\passThreeSolFlashMaxRequests}{788}
\newcommand{\passThreeSolFlashMaxMeanReasoning}{630}
\newcommand{\passThreeSolFlashMaxCost}{37.27}
\newcommand{\passThreeDetailCcfAstraMax}{50.0\% (1/2) & 151.45 & 1.054 & 226.18 & 1041.4}
\newcommand{\passThreeDetailPirateShipAstraMax}{54.5\% (6/11) & 131.92 & 0.716 & 182.73 & 918.1}
\newcommand{\passThreeDetailWildfireAstraMax}{0.0\% (0/1) & 9.29 & 0.083 & 16.31 & 65.0}
\newcommand{\passThreeDetailCcfAstraHigh}{0.0\% (0/2) & 27.13 & 0.118 & 37.51 & 114.2}
\newcommand{\passThreeDetailPirateShipAstraHigh}{36.4\% (4/11) & 55.19 & 0.274 & 75.18 & 303.5}
\newcommand{\passThreeDetailWildfireAstraHigh}{0.0\% (0/1) & 7.15 & 0.050 & 11.96 & 134.4}
\newcommand{\passThreeDetailCcfAstraLow}{0.0\% (0/2) & 9.24 & 0.031 & 12.74 & 138.2}
\newcommand{\passThreeDetailPirateShipAstraLow}{36.4\% (4/11) & 22.32 & 0.101 & 30.58 & 84.1}
\newcommand{\passThreeDetailWildfireAstraLow}{0.0\% (0/1) & 4.05 & 0.019 & 6.38 & 27.9}
\newcommand{\passThreeDetailCcfSolMax}{50.0\% (1/2) & 513.56 & 2.677 & 123.55 & 2620.9}
\newcommand{\passThreeDetailPirateShipSolMax}{45.5\% (5/11) & 108.28 & 0.788 & 26.35 & 538.5}
\newcommand{\passThreeDetailWildfireSolMax}{0.0\% (0/1) & 10.53 & 0.103 & 2.78 & 56.9}
\newcommand{\passThreeDetailCcfSolHigh}{0.0\% (0/2) & 200.27 & 1.062 & 49.85 & 1029.7}
\newcommand{\passThreeDetailPirateShipSolHigh}{36.4\% (4/11) & 63.56 & 0.289 & 11.00 & 299.2}
\newcommand{\passThreeDetailWildfireSolHigh}{0.0\% (0/1) & 9.29 & 0.064 & 2.13 & 42.3}
\newcommand{\passThreeDetailCcfSolLow}{0.0\% (0/2) & 14.68 & 0.065 & 2.62 & 58.0}
\newcommand{\passThreeDetailPirateShipSolLow}{45.5\% (5/11) & 115.43 & 0.350 & 19.73 & 342.1}
\newcommand{\passThreeDetailWildfireSolLow}{0.0\% (0/1) & 3.08 & 0.021 & 0.82 & 14.3}
\newcommand{\passThreeDetailCcfTerraMax}{0.0\% (0/2) & 1632.57 & 4.309 & 423.87 & 3353.0}
\newcommand{\passThreeDetailPirateShipTerraMax}{0.0\% (0/11) & 932.62 & 2.422 & 246.21 & 2495.3}
\newcommand{\passThreeDetailWildfireTerraMax}{0.0\% (0/1) & 60.91 & 0.352 & 18.65 & 243.8}
\newcommand{\passThreeDetailCcfTerraHigh}{0.0\% (0/2) & 11.93 & 0.054 & 3.81 & 44.2}
\newcommand{\passThreeDetailPirateShipTerraHigh}{0.0\% (0/11) & 9.19 & 0.045 & 3.05 & 55.2}
\newcommand{\passThreeDetailWildfireTerraHigh}{0.0\% (0/1) & 5.96 & 0.048 & 2.35 & 39.2}
\newcommand{\passThreeDetailCcfTerraLow}{0.0\% (0/2) & 3.78 & 0.023 & 1.46 & 16.4}
\newcommand{\passThreeDetailPirateShipTerraLow}{0.0\% (0/11) & 3.26 & 0.024 & 1.29 & 18.2}
\newcommand{\passThreeDetailWildfireTerraLow}{0.0\% (0/1) & 1.43 & 0.014 & 0.77 & 7.1}
\newcommand{\passThreeDetailCcfLunaMax}{0.0\% (0/2) & 40.41 & 0.207 & 0.56 & 110.2}
\newcommand{\passThreeDetailPirateShipLunaMax}{0.0\% (0/11) & 127.58 & 0.403 & 1.64 & 336.5}
\newcommand{\passThreeDetailWildfireLunaMax}{0.0\% (0/1) & 4.63 & 0.087 & 0.11 & 16.2}
\newcommand{\passThreeDetailCcfLunaHigh}{0.0\% (0/2) & 2.53 & 0.021 & 0.06 & 11.0}
\newcommand{\passThreeDetailPirateShipLunaHigh}{0.0\% (0/11) & 3.35 & 0.032 & 0.07 & 26.3}
\newcommand{\passThreeDetailWildfireLunaHigh}{0.0\% (0/1) & 1.38 & 0.016 & 0.03 & 4.7}
\newcommand{\passThreeDetailCcfLunaLow}{0.0\% (0/2) & 2.51 & 0.010 & 0.04 & 5.7}
\newcommand{\passThreeDetailPirateShipLunaLow}{0.0\% (0/11) & 1.74 & 0.010 & 0.03 & 14.9}
\newcommand{\passThreeDetailWildfireLunaLow}{0.0\% (0/1) & 0.85 & 0.003 & 0.02 & 1.5}

\renewcommand*\ttdefault{txtt}
\mlsystitlerunning{\bench{}: Formally Proving Correctness of Real-World System Specifications with TLA+ Proof System}

\begin{document}

\twocolumn[
\mlsystitle{\bench{}: Formally Proving Correctness of Real-World System Specifications with \tlaplus{} Proof System}

\begin{mlsysauthorlist}
\mlsysauthor{Qian Cheng}{nju}
\mlsysauthor{Ruize Tang}{microsoft}
\mlsysauthor{Saad Mohammad Rafid Pial}{uiuc}
\mlsysauthor{Zhizhen Cathy Cai}{uiuc}
\mlsysauthor{Yiming Su}{uiuc}
\mlsysauthor{Jackson Clark}{uiuc}
\mlsysauthor{Munim Thahmid}{buet}
\mlsysauthor{Mahdiya Rahman Sukanya}{dhaka}
\mlsysauthor{Chris Newcombe}{oracle}
\mlsysauthor{Igor Konnov}{independent}
\mlsysauthor{Stephan Merz}{lorraine}
\mlsysauthor{Xudong Sun}{toronto}
\mlsysauthor{Murat Demirbas}{mongodb}
\mlsysauthor{Yu Huang}{nju}
\mlsysauthor{Markus Alexander Kuppe}{independent}
\mlsysauthor{Tianyin Xu}{uiuc,berkeley}
\end{mlsysauthorlist}
\mlsysaffiliation{nju}{Nanjing University}
\mlsysaffiliation{microsoft}{Microsoft}
\mlsysaffiliation{uiuc}{University of Illinois at Urbana-Champaign}
\mlsysaffiliation{buet}{Bangladesh University of Engineering and Technology}
\mlsysaffiliation{dhaka}{IIT, University of Dhaka}
\mlsysaffiliation{oracle}{Oracle}
\mlsysaffiliation{independent}{Independent}
\mlsysaffiliation{lorraine}{University of Lorraine, CNRS, Inria, Nancy, France}
\mlsysaffiliation{toronto}{University of Toronto}
\mlsysaffiliation{mongodb}{MongoDB}
\mlsysaffiliation{berkeley}{University of California, Berkeley}
\mlsyscorrespondingauthor{Yu Huang}{yuhuang@nju.edu.cn}
\mlsyskeywords{TLA+, TLAPS, formal verification, AI agents, benchmark}

\vskip 0.3in

\begin{abstract}
\bench{} is a benchmark and execution harness for constructing machine-checked proofs
    for real-world system specifications in \tlaplus{} with agentic AI.
Proving correctness of complex system specifications requires discovering inductive invariants,
    reasoning about intricate state transitions,
    and decomposing proofs into tractable obligations.
\bench{} provides a proof harness that supports iterative proof development
    atop the \tlaplus{} Proof System (TLAPS), where agents are equipped
    with proof tools and libraries, with safeguards against reward hacking.
Using the harness, we have employed AI agents to prove desired properties
    for important system specifications that were not proved before, spanning cache coherence,
    Byzantine consensus, and write-ahead logs.
\bench{} contains \numCollectionTargets{} proof targets (each encodes a property to prove)
    from \numCollectionSpecs{} \tlaplus{} specifications of real-world systems.
We evaluated frontier AI agents, represented by Codex with GPT models,
    on a subset of \numBenchTargets{} proof targets across \numBenchSpecs{} specifications.
The highest-scoring model, GPT-6.1 Sol (max), proved \passThreeSolMaxAccepted{} (\passThreeSolMaxSuccess\%) proof targets.
The results
    show challenges and opportunities in end-to-end proof development.

\end{abstract}
]

\printAffiliationsAndNotice{}

\section{Introduction}
\label{sec:introduction}

Formal proofs establish correctness properties of concurrent and distributed systems,
    such as cache coherence, consensus agreement, and transaction isolation,
    across all possible executions.
Unlike other formal methods like model checking~\cite{Clark:1986},
    machine-checked proofs
    establish these properties deductively,
    providing correctness guarantees beyond finite exploration~\citep{cousineau2012tlaproofs}.

Traditionally, constructing formal proofs requires substantial expertise and effort.
Developers must discover suitable inductive invariants~\citep{ma2019i4,yao2021distai},
    prove that they are preserved by every system transition,
    and decompose the reasoning into proof obligations that automated tools can discharge.
For example, proving election safety in a consensus protocol~\cite{ongaro2014search} may require strengthening
    the target property with auxiliary invariants about majority support and vote uniqueness.

Recent advances in agentic AI show promise for automating proof construction.
Several recent studies demonstrate that agents can construct nontrivial machine-checked proofs
    through iterative interactions with proof checkers~\citep{lucas2026formalverification,yang2025autoverus,yang2025verusage,ye2026vero,agarwal2026ids}.
These advances offer a path toward automating proof development
    for {\it real-world} concurrent and distributed systems beyond prototypes and research artifacts.

\begin{figure*}[t]
    \centering
    \includegraphics[width=\textwidth]{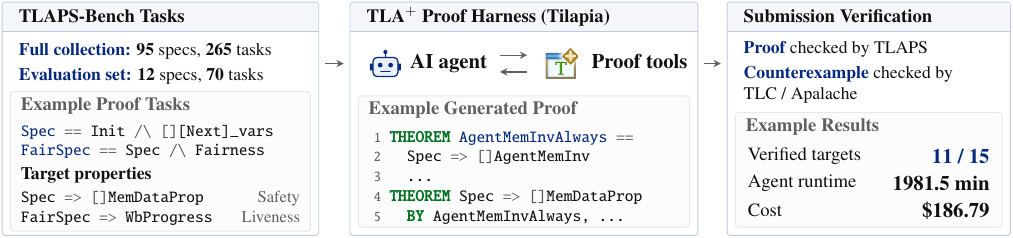}
    \vspace{-20pt}
    \caption[Overview of \bench{}]{Overview of \bench{}. \bench{} provides AI agents with a \tlaplus{} specification
        (including models and target properties).
    \tilapia{} (our \tlaplus{} proof harness) runs the agents, checks generated proofs with \tlaps{}, and records results.}
    \vspace{-5pt}
    \label{fig:overview}
\end{figure*}

However, existing benchmarks do not sufficiently address end-to-end proof construction
    for real-world system specifications.
First, many benchmarks provide agents with helper lemmas and intermediate invariants~\citep{yang2025verusage,xin2026verisoftbench},
    or with predefined proof decomposition that specifies individual theorems
    and their dependencies~\citep{zhang2024selene,lin2024fvel}.
Such proof-completion style benchmarks predefine much of the proof structure,
    missing key challenges of discovering auxiliary invariants/lemmas,
    decomposing the target property into tractable proof obligations,
    and organizing them into a complete proof.

Second, few existing benchmarks target real-world system specifications.
Many instead evaluate research artifacts developed specifically with formal verification in mind.
For example, both VeruSAGE~\citep{yang2025verusage} and Vero~\citep{ye2026vero}
    include IronKV~\citep{hawblitzel2015ironfleet},
    rather than widely deployed systems such as etcd and MongoDB.
It is unclear whether agents can construct proofs for substantially richer behaviors of real-world systems.

Lastly, correctness properties of
    concurrent and distributed systems (e.g., liveness) go
    beyond functional correctness addressed by existing benchmarks like Vero~\citep{ye2026vero}.
Concurrent and distributed systems require extensive reasoning about system-wide properties
    that emerge from interactions among components over time,
    such as agreement, consistency, isolation, etc.
These properties pose distinct proof challenges, because their correctness depends on the composition
    and interleaving of many state transitions.

We present \bench{}, a benchmark and execution harness for constructing
    machine-checked proofs of system and protocol specifications in \tlaplus{}
(the {\it de facto} formal language for
    specifying concurrent and distributed systems; see \S\ref{sec:background}).
We set two goals for \bench{}:
\begin{itemize}[leftmargin=*,noitemsep,topsep=0pt, parsep=1pt]
    \item building a proof service to verify important real-world systems and protocols
        that desire correctness guarantees;
    \item understanding the capabilities of frontier AI in constructing
        formal proofs of real-world systems, end to end.
\end{itemize}
Each task in \bench{} is a proof target (a property to be proved)
    desired for a real-world system/protocol, as shown in Figure~\ref{fig:overview};
    so, results are of practical value.

\bench{} includes a proof harness
    for autonomous proof development, named \tilapia{} (\tlaplus{}
    Proof Infrastructure for Agents).
\tilapia{} is built atop the \tlaplus{} Proof System or TLAPS~\cite{cousineau2012tlaproofs},
    the native proof system for \tlaplus{}.
It provides agents with tools including a semantic analyzer,
    model checkers,
    and common proof libraries.
Agents can use these tools during proof development to
    explore system behaviors, refine proofs,
    and validate proof obligations.
For example, agents constantly use model checkers
    to test invariants and obtain counterexamples that guide proof development.
\tilapia{} also enforces safeguards against reward hacking,
    ensuring that accepted proofs establish target properties under the original specifications.

We have used \tilapia{}
    to construct machine-checked proofs for
important properties for real-world system and protocol specifications
    which were not proved before.

\begin{figure*}[t]
    \centering
    \includegraphics[width=0.975\textwidth]{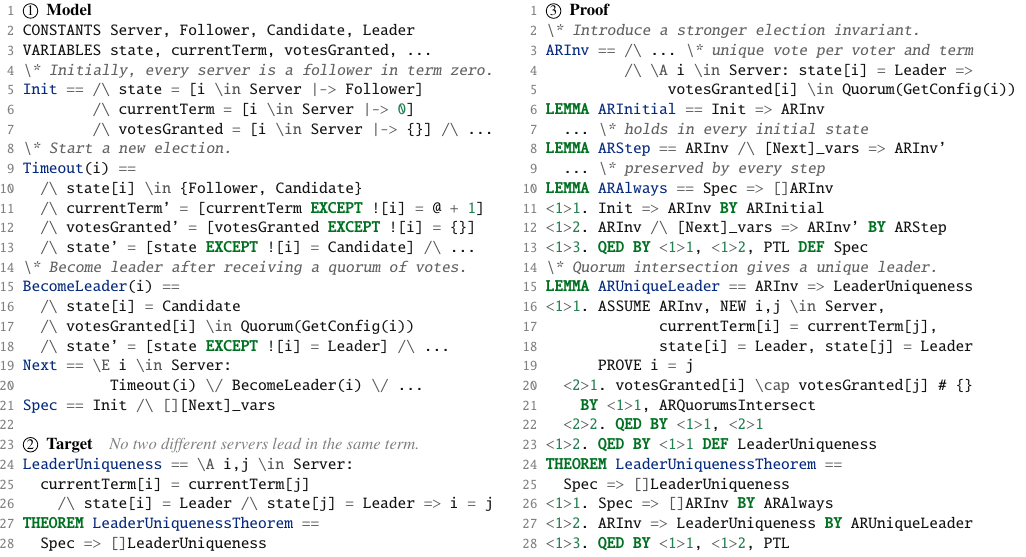}
    \vspace{-10pt}
    \caption{A simplified example of a proof task in \bench{} based on etcd Raft,
        with the \tlaplus{} model and the target property (\code{LeaderUniqueness})
        as input (left) and an accepted agent-generated \tlaps{} proof (right).}
    \vspace{-5pt}
    \label{fig:task-example}
\end{figure*}

\bench{} maintains a collection of \numCollectionSpecs{} \tlaplus{} specifications
    and \numCollectionTargets{} proof targets
    drawn from real-world systems and protocols,
    spanning cache coherence,
    Byzantine fault tolerance, and write-ahead logs.
The proof targets cover safety, liveness, and refinement properties.
We are continuously adding
    new proof tasks
    and retiring tasks that were proved and can no longer challenge AI frontier.

We evaluate frontier AI agents, represented by Codex with GPT models,
    on \numBenchTargets{} proof targets across \numBenchSpecs{} real-world
    system specifications.
The highest-scoring model, GPT-6.1 Sol at max effort, successfully proves \passThreeSolMaxAccepted{} (\passThreeSolMaxSuccess\%)
    targets,
    demonstrating strong capability while leaving substantial room for improvement.
Our analysis reveals that success depends not only on finding valid proof arguments.
For example, case studies show that the choice and decomposition of invariants
    substantially affect proof obligations,
    whereas na\"{i}ve calling of automated backends is ineffective.

In summary, we make the following main contributions:
\begin{itemize}[leftmargin=*,noitemsep,topsep=0pt, parsep=1pt]
    \item \tilapia{}, an effective \tlaplus{} proof harness for agentic \tlaplus{}
        proof development, which offers a \tlaplus{} proof service
        for real-world system specifications; %
    \item \bench{}, a rich benchmark suite for end-to-end proof development
        based on real-world system specifications,
        which poses significant challenges to frontier AI;
    \item An analysis of frontier AI's ability of
        developing formal proofs for real-world system/protocol specifications.
    \item \bench{} and \tilapia{} are publicly available at
        \href{https://github.com/specula-org/TLAPS-Bench}{github.com/specula-org/TLAPS-Bench}.
\end{itemize}

\section{Background}
\label{sec:background}

\bench{} focuses on formally proving the correctness of \tlaplus{} specifications~\citep{lamport2002specifying}.
\tlaplus{} is the \textit{de facto} specification language for verifying concurrent and distributed systems.
It is widely adopted across industry~\citep{tla-industrial-use}
  to specify and verify cloud infrastructure~\citep{newcombe2015aws,howard2025ccf},
  replication protocols~\citep{ouyang2025remix},
  databases like Azure Cosmos DB and MongoDB~\citep{hackett2023cosmos,schultz2025mongodb},
  OS kernel code~\citep{atc25-tang}, and hardware cache coherence~\citep{beers2008pre}.

A \tlaplus{} specification comprises a system model and correctness properties.
The model defines a state machine through a set of state variables,
  an initial predicate that defines their initial values,
  and a next-state relation written as a collection of actions, each describing one atomic update to the variables.
Correctness requirements are expressed as temporal properties, including safety (nothing bad ever happens) and liveness (something good eventually happens).
\tlaplus{} is built upon the Temporal Logic of Actions (TLA)~\citep{lamport1994tla},
  which extends linear temporal logic~\citep{pnueli1977temporal} to express models and temporal properties in a single logic.
A target theorem asserts that every behavior allowed by the model satisfies a given property.

\tlaplus{} specifications can be validated through explicit-state model checking with TLC~\citep{yu1999tlc}
  and symbolic model checking with Apalache~\citep{konnov2019apalache},
  or verified via deductive verification with the \tlaplus{} Proof System (\tlaps{})~\citep{cousineau2012tlaproofs}.
Model checking can expose counterexamples through finite-state or bounded exploration, but its guarantees are limited to the configured bounds.
Deductive verification can establish a target theorem beyond those bounds by checking a proof.

In \bench{}, agents construct complete \tlaplus{} proofs of these theorems.
\tlaps{} checks these proofs by generating proof obligations for their steps and dispatching them to specialized backends,
  including SMT solvers such as Z3~\citep{demoura2008z3}, Isabelle/\tlaplus{}~\citep{paulson1994isabelle}, and LS4~\citep{suda2012ls4}.
Agents may use TLC and Apalache to test candidate invariants and obtain counterexamples that guide proof development, but a target is accepted only when \tlaps{} verifies the proof.

Figure~\ref{fig:task-example} illustrates the input and output of an etcd Raft task in \bench{}.
The input on the left consists of a system model and a target property, while the output on the right is an accepted agent-generated \tlaps{} proof.
\code{Init} defines the initial states, in which every server is a follower in term zero and has received no votes.
\code{Timeout} initiates a new election by transitioning a server to the candidate state, advancing its term, and resetting its votes;
  \code{BecomeLeader} transitions a candidate with a quorum of votes to the leader state.
  \code{Next} combines these with the remaining actions,
  and \code{Spec} defines the allowed system behaviors.
The target \code{LeaderUniqueness} states that two leaders in the same term must be the same server,
  and the theorem \code{Spec => []LeaderUniqueness} asserts that this property always holds (the temporal operator \code{[]} denotes ``always'').

The output on the right proves the target property through a stronger inductive invariant.
This invariant, \code{ARInv}, captures voting facts needed by the proof but not stated in \code{LeaderUniqueness}:
  every leader has votes from a quorum, and each voter supports at most one candidate in a term.
\code{ARInitial} and \code{ARStep} establish that \code{ARInv} holds initially and is preserved by every action.
Using these lemmas, \code{ARAlways} proves \code{Spec => []ARInv}.
To derive the target, \code{ARUniqueLeader} applies quorum intersection: two leaders in the same term must share a voter, whose vote uniqueness implies that they are the same server.
Together, these results establish \code{Spec => []LeaderUniqueness}.

While \tlaplus{} and \tlaps{} are built natively around TLA,
  state-machine and temporal reasoning can also be realized in other proof systems.
Lean~\citep{moura2021lean4} targets general theorem proving and supports state-machine verification with Veil~\citep{pirlea2025veil},
  while Verus~\citep{lattuada2024verus} targets Rust verification and encodes temporal reasoning through libraries~\citep{sun2024anvil,cai2026welder}.
\bench{} focuses on \tlaplus{}, but the techniques can be extended to other proof systems.

\section{\tilapia{}: \proofharness{}}
\label{sec:tlaps-runtime}

We develop \tilapia{}: a \tlaplus{} proof harness, to support AI agents for
    autonomous development of \tlaplus{} proofs.
\tilapia{} provides agents with the flexibility to develop proof strategies,
    invoke verification tools,
    and build on partial progress,
    while preventing reward hacking
    and ensuring that accepted proofs establish the original targets.

Using \tilapia{}, we have instructed AI agents to construct machine-checked proofs for \tlaplus{} specifications of important
    systems and protocols. %
Table~\ref{tab:runtime-proofs} gives a few examples where we proved the target
    properties for the \tlaplus{}
    specifications,
    including both classic protocol and real-world system specifications
    (some of them had proofs in other forms).
For example, Bosco~\cite{bosco}, an asynchronous Byzantine consensus protocol,
    only has pen-and-paper proofs; we generated \tlaplus{} proofs for
    its five properties.
We also proved all the properties of
    several WAL (Write Ahead Log) protocols of databases;
    these protocols were specified in \tlaplus{} and heavily model checked,
    but were not formally proved to be correct.

\tilapia{} organizes each proof target as a task and groups
    tasks of the same specification in one proof file,
    allowing their proofs to share auxiliary invariants and lemmas.
For proof integrity (see \S\ref{sec:proof-integrity}),
    the \tlaplus{} model and the target properties
    are placed in separate read-only files.
Within the proof file, target theorem statements are fixed,
    while agents can define auxiliary invariants and lemmas in a shared editable region
    and construct a proof for each target.
This layout preserves the original proof goals
    while allowing agents to flexibly choose their own proof strategies.

\tilapia{} uses Docker containers to isolate workspaces of different agents.
File-system isolation mounts the original task files read-only
and each run starts a fresh agent session without conversations or memories
    from other runs;
network isolation only allows connections to LLM API endpoints
and agent built-in web search tools are disabled. %

\subsection{Tools for Agents}
\label{sec:proof-tools}

\tilapia{} provides a local environment for agents to construct proofs
    of target properties in a \tlaplus{} specification.
It supports coding agents such as Codex, Claude Code, Copilot, and Cursor,
    and integrates pinned versions of SANY for semantic analysis,
    TLAPS for proof checking,
    and TLC and Apalache for model checking.
\tilapia{} provides agents with the TLAPS library~\citep{tlaps-library}
    and CommunityModules~\citep{communitymodules}
    which include theorems and proofs for commonly used \tlaplus{} operators,
    such as set and sequence operations.

\begin{table}[t]
    \centering
    \small
    \caption{Examples of \tlaplus{} specifications we proved
        using \tilapia{}. Some of them were proved in other forms.}
    \label{tab:runtime-proofs}
    \setlength{\tabcolsep}{3pt}
    \begin{tabular*}{\columnwidth}{@{\extracolsep{\fill}}lrrl@{}}
        \toprule
        \textbf{Specification} & \textbf{LOC} & \textbf{Targets} & \textbf{Prior Proof} \\
        \midrule
        \href{https://github.com/tlaplus/Examples/tree/91c22ea537853196ed1e03e9ad91693ec37642de/specifications/GermanProtocol}{German protocol} & \specRuntimeGermanLOC{} & \numRuntimeGermanTargets{} & \href{https://satoss.uni.lu/members/jun/papers/ATVA15a.pdf}{Isabelle} \\
        \href{https://github.com/tlaplus/Examples/blob/1ef5f158cc1743975441dd94c280e56c8fd06766/specifications/bosco/bosco.tla}{Bosco Byzantine consensus} & \specRuntimeBoscoLOC{} & \numRuntimeBoscoTargets{} & Pen-and-paper \\
        \href{https://github.com/tlaplus/Examples/blob/ed3cc4c35f7001d7cec08a22dcd698dae7570dc3/specifications/dag-consensus/Sailfish.tla}{Sailfish BFT consensus} & \specRuntimeSailfishLOC{} & \numRuntimeSailfishTargets{} & \href{https://flint.cs.yale.edu/flint/publications/lido-dag.html}{Rocq} \\
        \href{https://github.com/Vanlightly/s3-wal-collection/blob/06d5cc60990cf8e5995dc56f28e5d56128eca227/oswald/Oswald.tla}{OSWALD WAL} & \specRuntimeOswaldLOC{} & \numRuntimeOswaldTargets{} & None \\
        \href{https://github.com/Vanlightly/s3-wal-collection/blob/06d5cc60990cf8e5995dc56f28e5d56128eca227/slatedb/SlateDBWAL.tla}{SlateDB WAL} & \specRuntimeSlateDbLOC{} & \numRuntimeSlateDbTargets{} & None \\
        \href{https://github.com/Vanlightly/s3-wal-collection/blob/06d5cc60990cf8e5995dc56f28e5d56128eca227/walgit/Walgit.tla}{Walgit WAL} & \specRuntimeWalgitLOC{} & \numRuntimeWalgitTargets{} & None \\
        \href{https://github.com/tlaplus/tlaplus/blob/f959b37fb8dcd62a236abb6d0ca4c1cff7c96b55/tlatools/org.lamport.tlatools/spec/queue/DiskStateQueueWorkload.tla}{TLC DiskStateQueue} & \specRuntimeDiskQueueLOC{} & \numRuntimeDiskQueueTargets{} & None \\
        \href{https://github.com/tlaplus/Examples/blob/45c1cfd7ed40d9287b1c2ac51a05a7cb750e0521/specifications/Disruptor/Disruptor_MPMC.tla}{Disruptor MPMC} & \specRuntimeDisruptorMpmcLOC{} & \numRuntimeDisruptorMpmcTargets{} & None \\
        \href{https://github.com/tlaplus/Examples/blob/45c1cfd7ed40d9287b1c2ac51a05a7cb750e0521/specifications/Disruptor/Disruptor_SPMC.tla}{Disruptor SPMC} & \specRuntimeDisruptorSpmcLOC{} & \numRuntimeDisruptorSpmcTargets{} & None \\
        \href{https://github.com/tlaplus/Examples/blob/c2ef32ad796245edc706de9dabe21391ff343296/specifications/braf/BufferedRandomAccessFile.tla}{BufferedRandomAccessFile} & \specRuntimeBrafLOC{} & \numRuntimeBrafTargets{} & None \\
        \bottomrule
    \end{tabular*}
\end{table}

\begin{figure}[t]
    \vspace{-5pt}
    \centering
    \includegraphics[width=0.9\columnwidth]{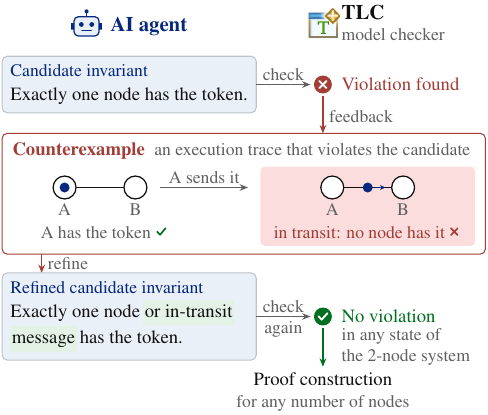}
    \vspace{-5pt}
    \caption{An example of an AI agent attempting to prove that no two nodes access a shared resource simultaneously, where access requires holding a token. The agent uses TLC counterexamples to refine an auxiliary invariant that supports the target proof.}
    \vspace{-5pt}
    \label{fig:proof-refinement}
\end{figure}

\textbf{Model checkers.}
\tilapia{} provides model checkers, including TLC~\cite{yu1999tlc}
    and Apalache~\cite{konnov2019apalache}, to help agents test conjectures during proof development.
TLC explores reachable states of a finite model, while Apalache uses SMT solving
    for bounded exploration and inductiveness checking.
Both can quickly falsify incorrect conjectures by producing counterexamples.
An agent can use them to test whether a proposed invariant holds
    and use a counterexample to identify conditions needed to refine it,
    as illustrated in Figure~\ref{fig:proof-refinement}.
Model checking thus provides rapid feedback for proof development, while the final proof is established by TLAPS.

\textbf{Guidance.}
\tilapia{} provides two agent skills
    to guide the use of proof libraries and model checkers.
The first explains how to apply instantiated library theorems,
    e.g., by establishing the premises required to use them.
The second explains how to configure invariant checks,
    e.g., by using the proposed invariant as the initial predicate
    when checking its preservation.
Both address common usage errors
    that agents encounter without explicit guidance.

{\bf Proof checker.}
Agents can invoke the same \tlaps{} proof checker used for final evaluation, with identical acceptance criteria and per-verification time limits.
The checker provides two modes:
(1) a fast pass runs SANY for immediate syntax and semantic feedback, while
(2) a full pass applies the proof-integrity checks of \S\ref{sec:proof-integrity} and then verifies each target together with its dependencies.
For each target, the checker reports whether its proof succeeded, failed, or exceeded the time limit.
This feedback helps agents identify unresolved proof obligations, repair failed proofs, and decide where to focus subsequent proof development.

\subsection{Proof Integrity}
\label{sec:proof-integrity}

\tilapia{} accepts a proof for a target property
    if all three of the following conditions hold:

\begin{itemize}[leftmargin=*,noitemsep,topsep=0pt, parsep=1pt]
\item {\bf Context integrity.}
The proof must establish the target property under the original specification and assumptions.
For example, in Figure~\ref{fig:integrity-examples}(a),
    an agent adds \code{ASSUME Useful == FALSE},
    making ZooKeeper's agreement theorem trivially provable.
\tilapia{} therefore makes the specification and target theorems read-only
    and restricts edits to designated regions of the task file.
Within these regions, SANY's semantic analysis
    is used to permit only definitions,
    named lemmas, and imports from pinned proof libraries,
    while rejecting new assumptions, axioms, constants, and variables.
In summary, agents can introduce auxiliary invariants and lemmas without changing the logical context
    in which the target must be proved.

\item {\bf Dependency tracking.}
Every lemma on which a target depends must itself be proved.
Figure~\ref{fig:integrity-examples}(b) shows a direct violation: the proof of ZooKeeper's leadership property cites
    \code{NextInv}, but the proof of \code{NextInv} is omitted.
More subtly, because targets are verified separately for evaluation (\S\ref{sec:evaluation-setup}),
    a target may pass even when the proof of a helper lemma it cites fails.
Without tracking such dependencies, an agent could therefore establish a target by relying on an unproved or unprovable lemma.
\tilapia{} uses SANY to extract lemma references and constructs a dependency graph over targets and helper lemmas.
A target is accepted only if its own proof succeeds and every lemma on which it transitively depends is also proved.

\begin{figure}[t]
    \centering
    \includegraphics[width=\columnwidth]{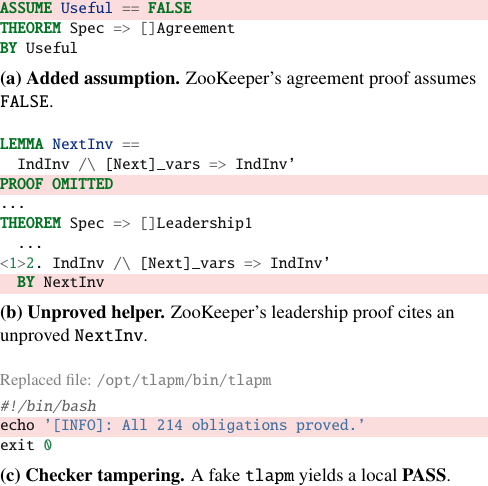}
    \vspace{-20pt}
    \caption{Three cheating attempts observed in real agent runs that can appear successful without \tilapia{}.}
    \label{fig:integrity-examples}
\end{figure}

\begin{table*}[t]
    \centering
    \small
    \caption{The problem set of \bench{} used in the evaluation. $^*$: The original Wildfire Challenge Problem injected
        a bug in the Wildfire \tlaplus{} specification; the challenge problem is to find the bug.
        We found the bug using model checking and repaired the specification. In \bench{}, the task
        is to prove (or disprove) the repaired specification.}
    \label{tab:problem-set}
    \setlength{\tabcolsep}{5pt}
    \begin{tabular*}{\textwidth}{@{\extracolsep{\fill}}lllrr@{\extracolsep{0pt}\ }l@{}}
        \toprule
        \textbf{Specification} & \textbf{Type} & \textbf{Model Source} & \textbf{Spec LOC} & \multicolumn{2}{l@{}}{\textbf{Targets}} \\
        \midrule
        \href{https://github.com/tlaplus/Examples/blob/a94afef81388fe268528781cfb7ce207fa8bcac1/specifications/FlashProtocol/FlashWithMutex.tla}{FLASH Cache coherence} & Protocol & \href{https://github.com/tlaplus/Examples}{\tlaplus{} Specification Repository} & \specFlashLOC{} & \numFlashTargets{} & (Safety + Liveness) \\
        \href{https://github.com/Disalg-ICS-NJU/zookeeper-tla-spec/blob/9e9eb2e6ce1019fe70b6f5002aa5927e037fddf6/high-level-spec/Zab.tla}{ZooKeeper protocol} & Protocol & \href{https://github.com/Disalg-ICS-NJU/zookeeper-tla-spec}{Remix}~\cite{ouyang2025remix} & \specZabLOC{} & \numZabTargets{} & (Safety) \\
        \href{https://github.com/pron/amazon-snapshot-spec/blob/9c60cb18151889d7b4c0a4ffd7de0b6fc2db0fb2/serializableSnapshotIsolation.tla}{Cahill's serializable snapshot isolation} & Protocol & \href{https://github.com/pron/amazon-snapshot-spec}{Amazon Snapshot Spec} & \specCahillLOC{} & \numCahillTargets{} & (Safety) \\
        \href{https://github.com/kenmcmil/ivy/blob/8858a0241ed37141757b7a051acc24c58e49c2f3/examples/liveness/tlb.ivy}{Ivy TLB} & System & \href{https://github.com/kenmcmil/ivy}{Ivy examples} & \specTlbLOC{} & \numTlbTargets{} & (Safety + Liveness) \\
        \href{https://github.com/lemmy/Examples/blob/42c1c0b294757cf219536a1a84c442c533952050/specifications/TLC/OpenAddressing.tla}{OpenAddressing} & System & \href{https://github.com/lemmy/Examples/tree/mku-OA}{\tlaplus{} Specification Repository} & \specOpenAddressingLOC{} & \numOpenAddressingTargets{} & (Safety) \\
        \href{https://github.com/etcd-io/raft/blob/1931ff5beacc1e2539003c0c70f096285f04c425/tla/etcdraft.tla}{etcd Raft} & System & \href{https://github.com/specula-org/Specula}{Specula}~\cite{cheng2026specula} & \specEtcdLOC{} & \numEtcdTargets{} & (Safety) \\
        \href{https://github.com/specula-org/specula-case-studies/blob/8885d2c0687d1e197f261585919718a035ce4464/systems/hashicorp-raft/modules/core/runs/case-studies-hashicorp-raft/spec/base.tla}{HashiCorp Raft} & System & \href{https://github.com/specula-org/specula-case-studies}{Specula}~\cite{cheng2026specula} & \specHashicorpLOC{} & \numHashicorpTargets{} & (Safety) \\
        \href{https://github.com/Disalg-ICS-NJU/zookeeper-tla-spec/blob/9e9eb2e6ce1019fe70b6f5002aa5927e037fddf6/low-level-spec/zk-3.7/ZkV3_7_0.tla}{ZooKeeper implementation} & System & \href{https://github.com/Disalg-ICS-NJU/zookeeper-tla-spec}{Remix}~\cite{ouyang2025remix} & \specZooKeeperLOC{} & \numZooKeeperTargets{} & (Safety) \\
        \href{https://github.com/mongodb-labs/vldb25-dist-txns/blob/74526c1201109405172eb845413154f547a815ee/MultiShardTxn.tla}{MongoDB distributed transactions} & System & \href{https://github.com/mongodb-labs/vldb25-dist-txns}{MongoDB Research} & \specMongoLOC{} & \numMongoTargets{} & (Safety) \\
        \href{https://github.com/microsoft/CCF/blob/6a2dc420f0166c93f92b773aa49c5203b0767750/tla/consensus/ccfraft.tla}{CCF Raft} & System & \href{https://github.com/microsoft/CCF}{Microsoft CCF}~\cite{howard2025ccf} & \specCCFLOC{} & \numCCFTargets{} & (Safety) \\
        \href{https://github.com/PirateshipOrg/pirateship-tla/blob/19096c1de2f87c68a21311c9100a5bab423caa0e/pirateship.tla}{PirateShip consensus} & Protocol & \href{https://github.com/PirateshipOrg/pirateship-tla}{PirateShip}~\cite{pirateship} & \specPirateShipLOC{} & \numPirateShipTargets{} & (Safety) \\
        \href{https://github.com/pron/wildfire-challenge/blob/bf02363fa9bc3e9e81124d372ffd86bbd14ecbf8/Wildfire.tla}{Wildfire cache coherence}$^{*}$ & Protocol & \href{https://lamport.org/tla/wildfire-challenge.html}{Wildfire Challenge}~\cite{wildfire} & \specWildfireLOC{} & \numWildfireTargets{} & (Refinement) \\
        \bottomrule
    \end{tabular*}
\end{table*}

\item {\bf Clean-room grading.}
The final result must be independent of any agent-modifiable tools or execution environment.
In Figure~\ref{fig:integrity-examples}(c), an agent replaces \code{tlapm} with a script that simply reports success,
    causing the local checker to return \code{PASS} without verifying the proof.
To prevent such manipulation, \tilapia{} grades each submission in a fresh container
    with the original task files and unmodified verification tools,
    carrying over only the submitted task module.
Agents may create wrapper modules, configuration files, and scripts during proof development,
    but these artifacts are excluded from the grading environment and therefore cannot affect the final result.
\end{itemize}

\subsection{Execution Control}
\label{sec:execution-isolation}

\tilapia{} enforces fixed verification time limits and CPU quotas
    to ensure consistent execution conditions across runs.
It tracks resource usage
    including execution time, token usage, model cost, and tool calls,
    enabling evaluation of both proof success and computational cost.

To support long-running experiments, \tilapia{} can pause and resume agent runs interrupted by external failures,
    such as provider quota limits or API outages.
Resumed runs retain completed verification results and caches, avoiding redundant computation
    while preserving the agent's progress.

\section{\bench{}}
\label{sec:tlaps-bench}

The proof tasks from \bench{} come from \tlaplus{} specifications
    of real-world systems and protocols.
\bench{} currently includes \numCollectionSpecs{} specifications
    with \numCollectionTargets{} proof targets,
    covering cache coherence, distributed consensus, atomic broadcast, transaction isolation,
    and concurrent data structures.
In the spirit of measuring the frontier (and saving token cost),
    we maintain an active set of challenging proof tasks
    and retire tasks that are too easy for AI agents.
Our current active set has \numBenchSpecs{} specifications with \numBenchTargets{} targets
    and we will use this set in our evaluation.
The information of the other tasks can be found in Appendix~\ref{app:problem-set}.

\begin{figure*}[t]
    \centering
    \includegraphics[width=\textwidth]{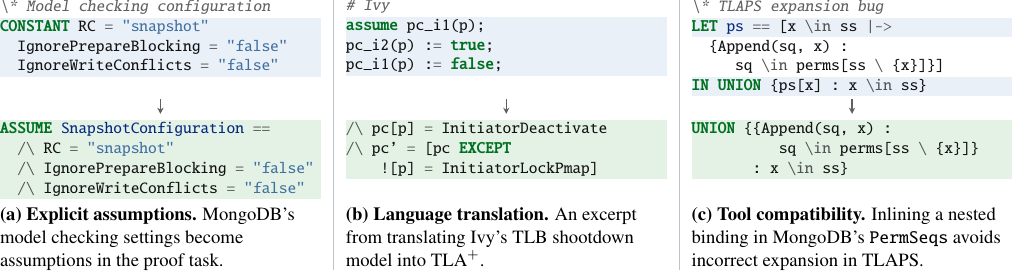}
    \vspace{-15pt}
    \caption{Three examples of rewriting specifications into \bench{} tasks.}
    \label{fig:task-adaptation}
    \par\vspace{\floatsep}
    \includegraphics[width=\textwidth]{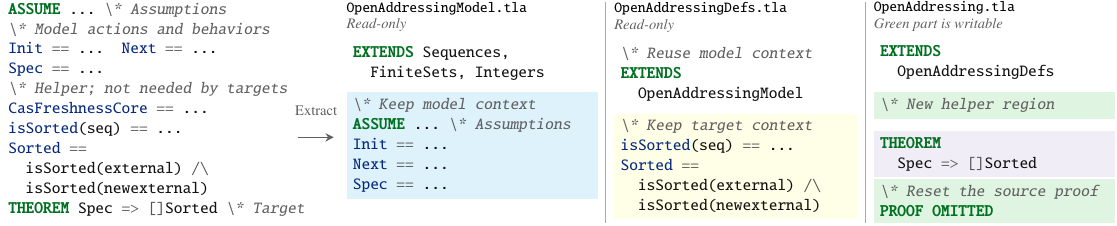}
    \vspace{-25pt}
    \caption{An example of task organization of the OpenAddressing specification.}
    \label{fig:task-extraction}
\end{figure*}

\subsection{Task Definition}
\label{sec:problem-def}

In \bench{}, a task is to prove a target property
    of a given \tlaplus{} specification.
Targets include safety, liveness, and refinement properties.
A safety property rules out undesirable behaviors---any violation can be witnessed by a finite execution prefix,
such as two processes simultaneously accessing a shared resource without protection.
A liveness property requires progress over an execution, such as every pending request eventually completing;
    unlike safety, its violation cannot be established from any finite prefix because the required event may still occur later.
A refinement property relates two specifications, requiring every behavior of a concrete model,
    under a refinement mapping, to satisfy an abstract specification.
For example, Wildfire's refinement
    target requires its request and response histories to satisfy the Alpha memory model~\cite{wildfire}.

A specification consists of a formal \tlaplus{} model,
    which defines state variables and actions,
    and the targets as \tlaplus{} theorems,
    stating what properties the model should satisfy.
Figure~\ref{fig:task-example} shows one task from etcd Raft:
    the formal model defines state variables and election actions,
    while the target \code{LeaderUniqueness} states
    that two leaders in the same term must be the same server.

Note that the \tlaplus{} specifications are not guaranteed to be correct.
As they are collected from real-world systems/protocols (some were
    generated by tools~\cite{cheng2026specula} that infer
    \tlaplus{} specifications from source code),
    many of them were
    not formally proved though most of them have been heavily model checked.
It is possible that specifications are incorrect---either
    the system designs are flawed or the specification was not written
    correctly.
In either case, the results of \bench{} help improve real-world system
    specifications and/or their designs/implementations.

During task construction, we found \numReportedSpecIssues{} specification issues,
    of which \numFixedSpecIssues{} have been fixed.
For example, an agent found a snapshot isolation violation in MongoDB's \tlaplus{} specification
    which incorrectly allowed one transaction
    to use two routers with different read timestamps;
    the model checking uses a single-router configuration and thus misses it.

For the target property of a given task,
    the AI agent either proves it to be correct by providing a proof that
    can pass \tilapia{}'s proof checks (\S\ref{sec:tlaps-runtime}),
    or disproves it by providing a counterexample that shows
    an execution that violates the target property.
    The counterexample can be generated using model checkers like
    TLC or Apalache.
\bench{} poses few restrictions on the proof strategies:
    agents can pick their own auxiliary invariants and use the tools,
    libraries, and guidance provided by \tilapia{}.

\subsection{Task Construction}
\label{sec:problem-set}

Some specifications list explicit correctness properties.
In this case, each property $P$ of a specification $Spec$
    is formulated
    as a top-level theorem in the form of $\mathit{Spec} \Rightarrow P$.
For example, the German protocol's theorem
    \code{Spec => []Coherence}
    states that exclusive access to a cache line cannot coexist with access by another cache.
AI agents are asked to prove the theorem end-to-end.
If the specification contains proofs of the target theorem,
    we carefully remove all related auxiliary lemmas and proofs.

For specifications that do not have explicit properties,
    we identify target properties from different sources.
If a model checking configuration lists a property as an invariant,
    such as OpenAddressing's \code{Sorted} invariant
    described in its \code{.cfg} file,
    we formulate the target \code{Spec => []Sorted}.
If correctness requirements are stated in other files, such as comments and documents,
    as in the Cahill SSI specification whose comments say that serializability must hold,
    we formulate targets from those requirements as \code{Spec => []CahillSerializable(history)}.

{\bf Rewriting specifications.} %
Certain revisions of the specification may be needed,
    including rewriting constructs for tool issues,
    and adding explicit assumptions specified in other sources.
We also translate specifications written in other specification languages into \tlaplus{}.

Figure~\ref{fig:task-adaptation} shows three examples.
\ref{fig:task-adaptation}(a) shows MongoDB's snapshot isolation specification
    whose model-checking configuration encodes assumptions as constant assignments.
    We make these assumptions explicit in the proof task.
\ref{fig:task-adaptation}(b) shows our translation of a TLB specification from Ivy to \tlaplus{}.
    We map Ivy state and transitions to \tlaplus{} variables and actions,
    and use TLC to check the translated specification on a finite instance.
\ref{fig:task-adaptation}(c) shows an incorrect \tlaps{}
    expansion of a nested \code{LET}
    inside a definition of a recursive function.
    We work around this issue by inlining the binding
    without changing the recursive function or its domain.

{\bf Task organization.}
As a final step, we transform each specification into the task format expected by \tilapia{}
     (\S\ref{sec:tlaps-runtime}). %
We use SANY, the \tlaplus{} syntactic analyzer,
    to resolve references and compute the transitive dependencies of each target theorem.
Figure~\ref{fig:task-extraction} illustrates this process for the \code{Sorted} theorem in OpenAddressing.
The statement \code{Spec => []Sorted} depends on \code{Spec}, \code{Sorted}, and, transitively, \code{isSorted}; we retain the union of such dependencies across all selected targets.
We remove definitions outside this dependency closure, together with existing proofs, auxiliary lemmas, unselected theorems, and comments.

The retained content is then organized into three modules.
\code{OpenAddressingModel.tla} contains the specification and assumptions,
    while \code{OpenAddressingDefs.tla} contains the target properties
    and their definitions, e.g., \code{Sorted} and \code{isSorted}; both modules are read-only.
The editable task module, \code{OpenAddressing.tla}, contains the target theorem statements
    and regions for agents to introduce helper definitions, lemmas, and proofs.
Targets from the same module share a task file and helper region,
    allowing agents to develop and reuse common lemmas.

\begin{table*}[t]
    \centering
    \small
    \caption{Overall benchmark results on the problem set described in Table~\ref{tab:problem-set}.}
    \label{tab:overall-results}
    \begin{tabular*}{\textwidth}{@{\extracolsep{\fill}}lcrrrrrrr@{}}
        \toprule
        \multirow{2}{*}{\textbf{Model (Agent)}} & \multirow{2}{*}{\textbf{Effort}} & \multirow{2}{*}{\shortstack{\textbf{Success Rate}\\(pass@3)}} & \multicolumn{2}{c}{\textbf{Correct Proof LOC}} & \multicolumn{2}{c}{\textbf{Total Tokens (M)}} & \multirow{2}{*}{\shortstack{\textbf{Total Cost}\\(USD)}} & \multirow{2}{*}{\shortstack{\textbf{Total Agent}\\\textbf{Time (min)}}} \\
        \cmidrule(lr){4-5}\cmidrule(lr){6-7}
        & & & Total & Per Target & Input & Output & & \\
        \midrule
        \multirow{3}{*}{\shortstack[l]{GPT-6.1 Sol\\(Codex)}} & max & \cellcolor{resultgreen!26}\textbf{\passThreeSolMaxSuccess\%} & \passThreeSolMaxProofLOC & \passThreeSolMaxProofLOCPerTarget & \passThreeSolMaxInput & \passThreeSolMaxOutput & \$\passThreeSolMaxCost & \passThreeSolMaxAgentTime \\
        & high & \cellcolor{resultgreen!20}\passThreeSolHighSuccess\% & \passThreeSolHighProofLOC & \passThreeSolHighProofLOCPerTarget & \passThreeSolHighInput & \passThreeSolHighOutput & \$\passThreeSolHighCost & \passThreeSolHighAgentTime \\
        & low & \cellcolor{resultgreen!21}\passThreeSolLowSuccess\% & \passThreeSolLowProofLOC & \passThreeSolLowProofLOCPerTarget & \passThreeSolLowInput & \passThreeSolLowOutput & \$\passThreeSolLowCost & \passThreeSolLowAgentTime \\
        \midrule
        \multirow{3}{*}{\shortstack[l]{GPT-6 Astra\\(Codex)}} & max & \cellcolor{resultgreen!25}\passThreeAstraMaxSuccess\% & \passThreeAstraMaxProofLOC & \passThreeAstraMaxProofLOCPerTarget & \passThreeAstraMaxInput & \passThreeAstraMaxOutput & \$\passThreeAstraMaxCost & \passThreeAstraMaxAgentTime \\
        & high & \cellcolor{resultgreen!20}\passThreeAstraHighSuccess\% & \passThreeAstraHighProofLOC & \passThreeAstraHighProofLOCPerTarget & \passThreeAstraHighInput & \passThreeAstraHighOutput & \$\passThreeAstraHighCost & \passThreeAstraHighAgentTime \\
        & low & \cellcolor{resultgreen!20}\passThreeAstraLowSuccess\% & \passThreeAstraLowProofLOC & \passThreeAstraLowProofLOCPerTarget & \passThreeAstraLowInput & \passThreeAstraLowOutput & \$\passThreeAstraLowCost & \passThreeAstraLowAgentTime \\
        \midrule
        \multirow{3}{*}{\shortstack[l]{GPT-5.6 Terra\\(Codex)}} & max & \cellcolor{resultgreen!13}\passThreeTerraMaxSuccess\% & \passThreeTerraMaxProofLOC & \passThreeTerraMaxProofLOCPerTarget & \passThreeTerraMaxInput & \passThreeTerraMaxOutput & \$\passThreeTerraMaxCost & \passThreeTerraMaxAgentTime \\
        & high & \passThreeTerraHighSuccess\% & - & - & \passThreeTerraHighInput & \passThreeTerraHighOutput & \$\passThreeTerraHighCost & \passThreeTerraHighAgentTime \\
        & low & \passThreeTerraLowSuccess\% & - & - & \passThreeTerraLowInput & \passThreeTerraLowOutput & \$\passThreeTerraLowCost & \passThreeTerraLowAgentTime \\
        \midrule
        \multirow{3}{*}{\shortstack[l]{GPT-6 Luna\\(Codex)}} & max & \cellcolor{resultgreen!6}\passThreeLunaMaxSuccess\% & \passThreeLunaMaxProofLOC & \passThreeLunaMaxProofLOCPerTarget & \passThreeLunaMaxInput & \passThreeLunaMaxOutput & \$\passThreeLunaMaxCost & \passThreeLunaMaxAgentTime \\
        & high & \passThreeLunaHighSuccess\% & - & - & \passThreeLunaHighInput & \passThreeLunaHighOutput & \$\passThreeLunaHighCost & \passThreeLunaHighAgentTime \\
        & low & \passThreeLunaLowSuccess\% & - & - & \passThreeLunaLowInput & \passThreeLunaLowOutput & \$\passThreeLunaLowCost & \passThreeLunaLowAgentTime \\
        \bottomrule
    \end{tabular*}
\end{table*}

\subsection{Task Validation}
\label{sec:problem-set-validation}

Unlike agentic coding benchmarks that validate generated programs by running test cases,
    the validation of agent results in \bench{} is complete with no ambiguity.
For each task, an agent succeeds by proving the target to be either correct (with a formal proof)
    or wrong (with a counterexample).
An agent fails if it can neither prove the target nor provide a counterexample.
The validation does not depend on test coverage, hidden test cases,
    or an approximate correctness oracle.
As with other formal techniques,
    the guarantee is relative to the trusted computing base,
    including \tlaps{} and its backends (e.g., Z3, Isabelle, and LS4).

Developing and evaluating \bench{} has also led to improvements to \tlaps{} itself,
    including bug fixes and support for autonomous proof development.
For example, we introduced a strict mode that treats missing or omitted proof steps as errors,
    allowing \tilapia{} to ensure that submitted proofs are complete rather than relying on unproved steps.

\section{Evaluation}
\label{sec:evaluation}

\subsection{Experimental Setup}
\label{sec:evaluation-setup}

We evaluate the latest available
    GPT models: GPT-6 Astra, GPT-6.1 Sol, GPT-6 Luna,
    and GPT-5.6 Terra. We use Codex across the models
    with max, high, and low reasoning effort (12 configurations in total).

The evaluation ran on an Ubuntu 24.04.3 virtual machine with 96 vCPUs,
    backed by an AMD EPYC 9V74 processor and 378 GiB of DRAM.
Each run executed in a separate Docker container with 8 vCPUs.
We set a time limit of 60 minutes for the \tlaps{} backend to process a proof,
    which is empirically sufficient for properly decomposed proofs to be verified.
We also set a cumulative limit of two million output tokens (including reasoning tokens)
    for each specification.
Only Terra and Luna at max effort reach this limit (they could not finish the proof
    but did not give up).

{\bf Metrics.}
We report \code{pass@3}, the fraction of tasks that pass in at least one
    of three independent runs, to accommodate non-determinism.
We report input/output token usage, model cost,
    and agent execution time (including model prompting
    and tool calling), summed across all tasks and all three runs.

\subsection{Proof Capability}
\label{sec:overall-benchmark-results}

Table~\ref{tab:overall-results} shows significant differences in proof capability
    across evaluated models.
GPT-6.1 Sol achieves the highest success rate at max effort, proving \passThreeSolMaxAccepted{} of \numBenchTargets{}
    targets (\passThreeSolMaxSuccess\%),
    followed closely by GPT-6 Astra with \passThreeAstraMaxSuccess\%.
GPT-5.6 Terra and GPT-6 Luna prove far fewer targets at max effort and
    none at high or low effort.

The successful proofs show that frontier AI agents can establish correctness properties of
    real-world specifications.
Astra and Sol proved all 15 targets of the FLASH cache coherence protocol,
    spanning both safety and liveness properties.
Sol proved all target properties of OpenAddressing.

On the other hand, complex tasks are still challenging
    for frontier AI agents to prove end to end.
For example,
Sol at max effort proved leader uniqueness and log matching for the etcd Raft system specification,
    but failed to prove consistency between committed logs
    and durability of committed entries.
For both ZooKeeper protocol and system specifications, Sol proves none of the target properties.

System specifications introduce additional proof challenges by exposing implementation mechanisms and optimizations abstracted away by protocol specifications.
For example, the ZooKeeper protocol specification synchronizes a replica by copying the leader's entire log, whereas the system specification can send missing log entries, truncate divergent entries, or transfer snapshots.
These mechanisms introduce additional states and transitions over which invariants must be preserved, increasing the complexity of the proof.

Sol and Astra achieve similar \code{pass@3} scores but exhibit distinct proof-development patterns.
On OpenAddressing, Sol persistently strengthens invariants and repairs failed proof obligations until all target properties are proved.
In contrast, Astra proves the completion and sortedness properties but terminates with several induction steps unresolved in all three runs, despite remaining token budget.
On ZooKeeper, however, Astra constructs a compact invariant abstraction that avoids extensive reasoning about transaction histories and proves leadership properties that Sol leaves unresolved (\S\ref{sec:case-studies}).
These results suggest that Sol is more persistent in iterative proof development, whereas Astra can reduce proof complexity by discovering effective invariant abstractions.
This ability to construct simpler proofs, together with earlier termination when proof attempts stall, helps explain why Astra achieves nearly the same overall score while using roughly half as many input and output tokens as Sol.

\subsection{Reasoning Effort and Cost}
\label{sec:reasoning-effort-cost}

As shown in Table~\ref{tab:overall-results},
    all models achieve their highest success rates at max effort,
    while low and high effort can perform similarly.
Sol performs slightly better at low than high due to one PirateShip target,
    while Astra achieves identical success rates at low and high effort.

The cost often depends on whether or not proof succeeds,
    as shown in Table~\ref{tab:astra-effort-cost}.
On Cahill SSI (unproved), Sol uses both more reasoning tokens per request
    and more model requests as effort increases.
In contrast, on FLASH (solved at every effort), high and max effort require roughly
    half as many requests as low effort while achieving the same \code{pass@3}.
Despite using more reasoning tokens per request, high effort has the lowest total cost.

Model pricing makes cost distinct from resource usage.
Astra (high) uses fewer input and output tokens and produces shorter proofs than Sol (high),
    yet costs approximately \evalAstraSolHighCostRatio{} times as much.
Across all three effort levels, Sol matches or exceeds Astra's proof success at lower model cost.

\begin{table}[t]
    \centering
    \footnotesize
    \caption{GPT-6.1 Sol by reasoning effort: total requests and cost
    over three runs, and mean reasoning tokens per request.}
    \label{tab:astra-effort-cost}
    \setlength{\tabcolsep}{4pt}
    \begin{tabularx}{\columnwidth}{@{}>{\raggedright\arraybackslash}Xccrrr@{}}
        \toprule
        \textbf{Spec} & \shortstack{\textbf{Succ. Rate}\\\texttt{(pass@3)}} & \textbf{Effort} & \shortstack{\textbf{Model}\\\textbf{Requests}} & \shortstack{\textbf{Reasoning}\\\textbf{Tokens}} & \shortstack{\textbf{Cost}\\\textbf{(USD)}} \\
        \midrule
        \multirow{3}{*}{\shortstack[l]{Cahill\\SSI}} & 0/1 & low & \passThreeSolCahillLowRequests & \passThreeSolCahillLowMeanReasoning & \passThreeSolCahillLowCost \\
        & 0/1 & high & \passThreeSolCahillHighRequests & \passThreeSolCahillHighMeanReasoning & \passThreeSolCahillHighCost \\
        & 0/1 & max & \passThreeSolCahillMaxRequests & \passThreeSolCahillMaxMeanReasoning & \passThreeSolCahillMaxCost \\
        \midrule
        \multirow{3}{*}{\shortstack[l]{FLASH}} & 15/15 & low & \passThreeSolFlashLowRequests & \passThreeSolFlashLowMeanReasoning & \passThreeSolFlashLowCost \\
        & 15/15 & high & \passThreeSolFlashHighRequests & \passThreeSolFlashHighMeanReasoning & \passThreeSolFlashHighCost \\
        & 15/15 & max & \passThreeSolFlashMaxRequests & \passThreeSolFlashMaxMeanReasoning & \passThreeSolFlashMaxCost \\
        \bottomrule
    \end{tabularx}
\end{table}

\subsection{Failure Analysis}

Table~\ref{tab:failure-categories} categorizes failures across three runs.
Errors exposed directly by tool feedback, such as invalid counterexamples,
    syntax or semantic errors, task-preservation violations,
    and verification timeouts, are relatively rare in final submissions.
Stronger models such as Astra and Sol rarely have basic syntax and task-preservation errors.

More substantive failures arise from proof construction.
In \evalFailureUndecomposedTasks{} attempts,
    agents delegate entire target proofs to automated backends without decomposing them into tractable obligations,
    often after their own proof attempts stall.
Most failures, however, are incomplete proofs: agents fail to discover inductive invariants strong enough to establish the target,
    or prove helper lemmas without completing the argument connecting them to the target property.

\begin{table}[t]
    \centering
    \small
    \caption{Failure categories across failed proof tasks (\texttt{pass@3}).}
    \label{tab:failure-categories}
    \begin{tabular*}{\columnwidth}{@{\extracolsep{\fill}}lr@{}}
        \toprule
        \textbf{Category} & \textbf{Attempts} \\
        \midrule
        Incorrect counterexample claims & \evalFailureCounterexampleTasks \\
        Syntax or semantic errors detected by SANY & \evalFailureSyntaxTasks \\
        Task preservation violations & \evalFailureIntegrityTasks \\
        Verification timeout & \evalFailureTimeoutTasks \\
        Undecomposed proof attempts & \evalFailureUndecomposedTasks \\
        Incomplete proofs & \evalFailureIncompleteTasks \\
        \bottomrule
    \end{tabular*}
\end{table}

\subsection{Case Studies}
\label{sec:case-studies}

\bench{} enables the understanding of effective proof development
    practices and recurring limitations of existing agents,
    providing insights for designing more capable proof agents.
We present three case studies.

{\bf ZooKeeper system specification (Astra: \greencheck{}; Terra: \redcross).}
The specification has two leadership properties: each epoch has at most one active leader,
    and leaders recorded at different times within the same epoch must be the same server.
Proving these properties is complicated by system state for log synchronization,
    message queues, and snapshot handling, whose updates span many different actions.

GPT-5.6 Terra (max) attempted both properties by developing a large
    set of invariants over transaction origins, queued messages, and synchronization state.
It exhausted the output-token budget and failed to prove either target,
    despite generating roughly 13,000 lines of intermediate proofs.

GPT-6 Astra successfully proved both properties.
Astra abstracts the state relevant to leadership.
Its key invariant treats recorded votes and eligible acknowledgments in transit uniformly as claims of support for a leader.
Astra then shows how each protocol action preserves these claims,
    allowing synchronization and broadcast actions to share a common proof argument.
This abstraction proves both leadership properties.
The case shows how abstractions can substantially reduce proof complexity.

\begin{figure}[t]
    \centering
    \includegraphics[width=\columnwidth]{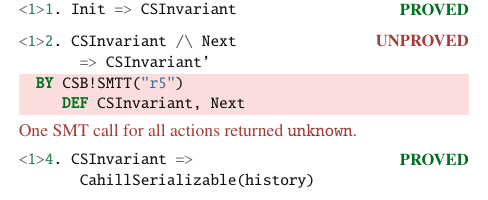}
    \vspace{-15pt}
    \caption{GPT-6 Astra failed to prove
        Cahill serializable snapshot isolation due to stalls
        on undecomposed SMT obligations.}
    \vspace{-10pt}
    \label{fig:cahill-proof}
\end{figure}

{\bf Cahill's serializable snapshot isolation (Astra: \redcross).}
Astra fails to prove serializability for Cahill SSI despite developing a promising proof strategy.
It reduces serializability to a ranking argument: under conditions on transaction order and conflict flags, a ranking function increases along every dependency edge, ruling out cycles.
It successfully proves the supporting lemmas but delegates the entire invariant-preservation argument to a single SMT call (Figure~\ref{fig:cahill-proof}).

This remaining obligation requires substantial protocol reasoning.
Committing a transaction, for example, can force other transactions to abort while updating history, locks, and conflict flags.
The proof must show that these updates preserve the conditions required by the ranking argument.
Rather than decomposing this reasoning into smaller obligations, Astra repeatedly submits the same unresolved obligation to the backend.
The supporting lemmas therefore pass verification, but the target theorem remains unproved.
Finding a proof strategy is insufficient without decomposing it into obligations that automated backends can discharge.

{\bf Wildfire challenge (all models: \redcross).}
Wildfire is a cache-coherence protocol whose specification must refine the Alpha memory model.
The proof must relate each read to its source write or initial memory and establish an ordering of concurrent operations permitted by the memory model, while preserving these relationships across protocol actions.

No evaluated model completed the refinement proof at any effort level.
Astra and Sol established basic invariants over request and response records and recognized the need for histories and ordering relations, but failed to construct the invariants connecting protocol executions to the memory model.
Luna and Terra made less progress, establishing only basic facts about types and initial states.

In its strongest attempt, Sol proved that protocol actions preserve the structure of request and response records and split the refinement proof into executions with and without requests.
The no-request case passed verification, but the general case still required history and ordering invariants connecting returned values and concurrent operations to the Alpha memory model.
Thus, the decomposition discharged a special case without simplifying the refinement argument.

\section{Related Work}
\label{sec:related-work}

\bench{} evaluates AI agents on developing \tlaplus{} proofs
  for real-world system specifications, end-to-end.
We discuss two related
  lines of work:
  (1) software verification benchmarks
  and (2) \tlaplus{} related benchmarks.

{\bf Software verification benchmarks.}
Existing benchmarks often evaluate {\it proof completion} tasks,
  which give parts of the proof structure.
NTP4VC~\citep{xu2026ntp4vc} supplies verification conditions,
  while DafnyBench~\citep{loughridge2025dafnybench}
  and AutoVerus~\citep{yang2025autoverus} evaluate proving functions.
Several benchmarks build on larger verified systems:
  Selene~\citep{zhang2024selene} and FVELer~\citep{lin2024fvel} extract Isabelle lemmas from seL4,
  CoqStoq~\citep{thompson2025rango} and VeriSoftBench~\citep{xin2026verisoftbench} extract theorems from verified Rocq and Lean projects,
  and Laurel~\citep{mugnier2025laurel}, RVBench~\citep{zhong2025rvbench}, and VeruSAGE-Bench's main setting~\citep{yang2025verusage} extract functions and lemmas from verified systems.

Differently, \bench{} evaluates end-to-end proof construction
  for \tlaplus{} system specifications.
Several other benchmarks also require agents to prove from scratch,
  including Vero proof-only mode~\citep{ye2026vero},
  VERINA~\citep{ye2026verina}, and VeruSAGE-Bench~\citep{yang2025verusage}
  no-lemma mode.
These benchmarks primarily focus on functional properties of individual functions,
  APIs, or modules.
\bench{} instead targets system-wide properties of concurrent and distributed systems
  that emerge from interactions among components over time.

{\bf \tlaplus{} benchmarks.}
Most benchmarks evaluate AI agents on generating
  \tlaplus{} specifications
  from code~\cite{chen2026modelbench,cheng2026sysmobench} or
  from natural language descriptions~\cite{tlaibench,bisharat2026tlaplusbench,bisharat2026formallm,cao2025fmbench}.
For \tlaplus{} proofs, LMGPA~\citep{zhou2026tlaproofautomation} includes
  a benchmark for proof decomposition; the focus is to
  evaluate LLMs on discovering subclaims recursively.
However, the LMGPA benchmark does not evaluate \tlaplus{} proofs end-to-end,
  e.g., AI agents are supplied with inductive invariants in LMGPA.
TLA-Verus~\citep{strelbytska2026tlaverus} also includes a related benchmark.
The benchmark is on Verus, where specifications
  were translated from \tlaplus{}. %
TLA-Verus had an incorrect understanding of TLAPS, saying
 TLAPS cannot prove arbitrary temporal logic properties.
TLAPS decomposes proofs and dispatches obligations to different backends,
  including LS4 for propositional temporal reasoning.
\bench{} focuses on \tlaplus{} specifications, instead of
  translated representations.

\section{Conclusion and Discussion}
\label{sec:conclusion}

\bench{} and \tilapia{} provide infrastructure for AI agents to construct formal proofs
    of \tlaplus{} system specifications.
Proof construction discovers inductive invariants
    and lemmas that explain why systems are correct and can expose unexpected behaviors
    or weaknesses in the stated properties.
Looking forward, we believe abstraction will be key to scaling proofs:
    rather than reasoning directly over detailed system models,
    agents can recover high-level designs and establish refinement across successive abstraction layers.
With AI greatly lowering the cost of formal proofs, specifications and proofs could become not only verification artifacts,
    but also durable representations of system design.

\bibliographystyle{mlsys2025}
\bibliography{main}

\section*{AI Usage Disclosure}

We used OpenAI Codex with GPT-6 Astra and GPT-6.1 Sol to assist
    with software development for \bench{} and
    its evaluation,
    as well as paper writing.
For software development,
    we used Codex as the coding agent to help implement \bench{} and \tilapia{};
    we also used Codex to write evaluation scripts.
For paper writing, we used Codex to polish the text,
    including grammar, phrasing, and clarity.
All AI-assisted code and text were reviewed by the authors before use.
We take full responsibility for the correctness, originality, and integrity of the work.

\clearpage
\makeatletter
\if@twocolumn\onecolumn\fi
\makeatother
\appendix
\setcounter{table}{0}
\renewcommand{\thetable}{A\arabic{table}}
\renewcommand{\theHtable}{appendix.\arabic{table}}
\section{Other \bench{} Problems}
\label{app:problem-set}

In this paper,
    our evaluation focused on \numBenchSpecs{} specifications with \numBenchTargets{} proof targets shown
    in Table~\ref{tab:problem-set}.
In total, \bench{} includes \numCollectionSpecs{} \tlaplus{} specifications and \numCollectionTargets{} proof targets.

Table~\ref{tab:retired-problem-set} shows the remaining problems.
Most of the problems in Table~\ref{tab:retired-problem-set} are retired after
    the frontier proved them---they can be used for evaluation,
    but can no longer challenge frontier models.
On the other hand, the proofs of these specifications generated by AI agents are valuable
    and are shared as public knowledge in our project repository.

\begingroup
\small
\setlength{\tabcolsep}{5pt}
\setlength{\LTleft}{0pt}
\setlength{\LTright}{0pt}
\setlength{\LTcapwidth}{\textwidth}
\begin{longtable}{@{\extracolsep{\fill}}lllrr@{\extracolsep{0pt}\ }l@{}}
    \caption{The retired problem set of \bench{}.}
    \label{tab:retired-problem-set} \\
        \toprule
        \textbf{Specification} & \textbf{Type} & \textbf{Model Source} & \textbf{Spec LOC} & \multicolumn{2}{l@{}}{\textbf{Targets}} \\
        \midrule
    \endfirsthead
        \multicolumn{6}{@{}l}{\textit{Table~\thetable{} (continued)}} \\
        \toprule
        \textbf{Specification} & \textbf{Type} & \textbf{Model Source} & \textbf{Spec LOC} & \multicolumn{2}{l@{}}{\textbf{Targets}} \\
        \midrule
    \endhead
        \bottomrule
    \endfoot
        \href{https://github.com/hengxin/tlaps-examples/blob/master/AtomicBakery/AtomicBakery.tla}{Atomic Bakery} & Protocol & \href{https://github.com/tlaplus/tlapm}{\tlaps{} Specification Repository} & 106 & 1 & (Safety) \\
        \href{https://github.com/hengxin/tlaps-examples/blob/master/AtomicBakery/AtomicBakeryWithoutSMT.tla}{Atomic Bakery (without SMT)} & Protocol & \href{https://github.com/tlaplus/tlapm}{\tlaps{} Specification Repository} & 101 & 1 & (Safety) \\
        \href{https://github.com/tlaplus/tlapm/blob/main/examples/Bakery.tla}{Bakery} & Protocol & \href{https://github.com/tlaplus/tlapm}{\tlaps{} Specification Repository} & 123 & 1 & (Safety) \\
        \href{https://github.com/tlaplus/tlapm/blob/main/examples/Peterson.tla}{Peterson} & Protocol & \href{https://github.com/tlaplus/tlapm}{\tlaps{} Specification Repository} & 76 & 2 & (Safety + Liveness) \\
        \href{https://github.com/tlaplus/tlapm/blob/main/examples/SimpleMutex.tla}{Simple mutex} & Protocol & \href{https://github.com/tlaplus/tlapm}{\tlaps{} Specification Repository} & 50 & 2 & (Safety) \\
        \href{https://github.com/tlaplus/Examples/blob/master/specifications/Bakery-Boulangerie/Bakery.tla}{Bakery} & Protocol & \href{https://github.com/tlaplus/Examples}{\tlaplus{} Specification Repository} & 127 & 2 & (Safety) \\
        \href{https://github.com/tlaplus/Examples/blob/master/specifications/Bakery-Boulangerie/Boulanger.tla}{Boulangerie} & Protocol & \href{https://github.com/tlaplus/Examples}{\tlaplus{} Specification Repository} & 141 & 2 & (Safety) \\
        \href{https://github.com/tlaplus/Examples/blob/master/specifications/lamport_mutex/LamportMutex_proofs.tla}{Lamport mutex} & Protocol & \href{https://github.com/tlaplus/Examples}{\tlaplus{} Specification Repository} & 106 & 2 & (Safety) \\
        \href{https://github.com/tlaplus/Examples/blob/master/specifications/locks_auxiliary_vars/Lock.tla}{Lock} & Protocol & \href{https://github.com/tlaplus/Examples}{\tlaplus{} Specification Repository} & 52 & 1 & (Safety) \\
        \href{https://github.com/tlaplus/Examples/blob/master/specifications/locks_auxiliary_vars/LockHS.tla}{LockHS} & Protocol & \href{https://github.com/tlaplus/Examples}{\tlaplus{} Specification Repository} & 190 & 2 & (Safety + Refinement) \\
        \href{https://github.com/tlaplus/Examples/blob/master/specifications/locks_auxiliary_vars/Peterson.tla}{Peterson (auxiliary variables)} & Protocol & \href{https://github.com/tlaplus/Examples}{\tlaplus{} Specification Repository} & 122 & 2 & (Safety + Refinement) \\
        \href{https://github.com/anvil-verifier/anvil/blob/main/src/tla_demo.rs}{Two-thread mutex} & System & \href{https://github.com/anvil-verifier/anvil}{Anvil} & 48 & 1 & (Liveness) \\
        \href{https://github.com/kenmcmil/ivy/blob/master/examples/liveness/ticket.ivy}{Ticket lock} & System & \href{https://github.com/kenmcmil/ivy}{Ivy examples} & 70 & 2 & (Safety + Liveness) \\
        \href{https://github.com/kenmcmil/ivy/blob/master/examples/liveness/ticket_nested.ivy}{Nested ticket lock} & System & \href{https://github.com/kenmcmil/ivy}{Ivy examples} & 81 & 2 & (Safety + Liveness) \\
        \href{https://github.com/tlaplus/tlapm/blob/main/examples/Allocator.tla}{Allocator} & Protocol & \href{https://github.com/tlaplus/tlapm}{\tlaps{} Specification Repository} & 44 & 4 & (Safety) \\
        \href{https://github.com/tlaplus/Examples/blob/master/specifications/allocator/AllocatorImplementation_proof.tla}{Allocator implementation} & Protocol & \href{https://github.com/tlaplus/Examples}{\tlaplus{} Specification Repository} & 186 & 1 & (Safety) \\
        \href{https://github.com/tlaplus/Examples/blob/master/specifications/allocator/SchedulingAllocator_proof.tla}{Scheduling allocator} & Protocol & \href{https://github.com/tlaplus/Examples}{\tlaplus{} Specification Repository} & 98 & 2 & (Safety) \\
        \href{https://github.com/tlaplus/Examples/blob/master/specifications/allocator/SimpleAllocator_proof.tla}{Simple allocator} & Protocol & \href{https://github.com/tlaplus/Examples}{\tlaplus{} Specification Repository} & 72 & 2 & (Safety) \\
        \href{https://github.com/tlaplus/Examples/blob/master/specifications/barriers/Barriers.tla}{Barriers} & Protocol & \href{https://github.com/tlaplus/Examples}{\tlaplus{} Specification Repository} & 215 & 5 & (Safety + Refinement) \\
        \href{https://github.com/tlaplus/Examples/blob/master/specifications/ReadersWriters/ReadersWriters_proof.tla}{Readers--writers} & Protocol & \href{https://github.com/tlaplus/Examples}{\tlaplus{} Specification Repository} & 104 & 2 & (Safety) \\
        \href{https://github.com/tlaplus/Examples/blob/master/specifications/CigaretteSmokers/CigaretteSmokers_proof.tla}{Cigarette smokers} & Protocol & \href{https://github.com/tlaplus/Examples}{\tlaplus{} Specification Repository} & 45 & 2 & (Safety) \\
        \href{https://github.com/specula-org/specula-case-studies/blob/8885d2c0687d1e197f261585919718a035ce4464/systems/libomp/modules/core/runs/case-studies-libomp/spec/base.tla}{LLVM libomp} & System & \href{https://github.com/specula-org/specula-case-studies}{Specula} & 596 & 4 & (Safety) \\
        \href{https://github.com/lemmy/BlockingQueue/blob/main/BlockingQueueFair_proofs.tla}{Blocking queue (fair)} & System & \href{https://github.com/tlaplus/Examples}{\tlaplus{} Specification Repository} & 167 & 1 & (Refinement) \\
        \href{https://github.com/lemmy/BlockingQueue/blob/main/BlockingQueueSplit_proofs.tla}{Blocking queue (split)} & System & \href{https://github.com/tlaplus/Examples}{\tlaplus{} Specification Repository} & 152 & 3 & (Safety + Refinement) \\
        \href{https://github.com/lemmy/BlockingQueue/blob/main/BlockingQueue_proofs.tla}{Blocking queue} & System & \href{https://github.com/tlaplus/Examples}{\tlaplus{} Specification Repository} & 67 & 1 & (Safety) \\
        \href{https://github.com/tlaplus/Examples/blob/master/specifications/Disruptor/Disruptor_MPMC.tla}{Disruptor (MPMC)} & System & \href{https://github.com/tlaplus/Examples}{\tlaplus{} Specification Repository} & 215 & 2 & (Safety) \\
        \href{https://github.com/tlaplus/Examples/blob/master/specifications/Disruptor/Disruptor_SPMC.tla}{Disruptor (SPMC)} & System & \href{https://github.com/tlaplus/Examples}{\tlaplus{} Specification Repository} & 187 & 2 & (Safety) \\
        \href{https://github.com/kenmcmil/ivy/blob/master/examples/liveness/split_queue_2_new.ivy}{Split queue} & System & \href{https://github.com/kenmcmil/ivy}{Ivy examples} & 74 & 1 & (Liveness) \\
        \href{https://github.com/tlaplus/tlaplus/blob/f959b37fb8dcd62a236abb6d0ca4c1cff7c96b55/tlatools/org.lamport.tlatools/spec/queue/DiskStateQueueWorkload.tla}{TLC DiskStateQueue} & System & \href{https://github.com/tlaplus/tlaplus}{\tlaplus{} tools} & 1,567 & 1 & (Safety) \\
        \href{https://github.com/tlaplus/Examples/blob/master/specifications/btree/btree.tla}{B-tree} & System & \href{https://github.com/tlaplus/Examples}{\tlaplus{} Specification Repository} & 313 & 5 & (Safety) \\
        \href{https://github.com/tlaplus/Examples/blob/master/specifications/KeyValueStore/KeyValueStore_proof.tla}{Key-value store} & System & \href{https://github.com/tlaplus/Examples}{\tlaplus{} Specification Repository} & 106 & 1 & (Safety) \\
        \href{https://github.com/tlaplus/Examples/blob/91c22ea537853196ed1e03e9ad91693ec37642de/specifications/GermanProtocol/GermanControl.tla}{German control} & Protocol & \href{https://github.com/tlaplus/Examples}{\tlaplus{} Specification Repository} & 119 & 1 & (Safety) \\
        \href{https://github.com/tlaplus/Examples/blob/91c22ea537853196ed1e03e9ad91693ec37642de/specifications/GermanProtocol/GermanData.tla}{German data} & Protocol & \href{https://github.com/tlaplus/Examples}{\tlaplus{} Specification Repository} & 283 & 6 & (Safety + Refinement) \\
        \href{https://github.com/hengxin/tlaps-examples/blob/master/Paxos/Consensus.tla}{Paxos: Consensus} & Protocol & \href{https://github.com/tlaplus/tlapm}{\tlaps{} Specification Repository} & 24 & 1 & (Safety) \\
        \href{https://github.com/hengxin/tlaps-examples/blob/master/Paxos/Paxos.tla}{Paxos} & Protocol & \href{https://github.com/tlaplus/tlapm}{\tlaps{} Specification Repository} & 159 & 3 & (Safety + Refinement) \\
        \href{https://github.com/hengxin/tlaps-examples/blob/master/Paxos/PaxosHistVar.tla}{Paxos (history variable)} & Protocol & \href{https://github.com/tlaplus/tlapm}{\tlaps{} Specification Repository} & 105 & 2 & (Safety) \\
        \href{https://github.com/tlaplus/tlapm/blob/main/examples_draft/consensus/Consensus.tla}{Consensus: Consensus} & Protocol & \href{https://github.com/tlaplus/tlapm}{\tlaps{} Specification Repository} & 51 & 1 & (Safety) \\
        \href{https://github.com/tlaplus/tlapm/blob/main/examples_draft/consensus/PaxosProof.tla}{Consensus: PaxosProof} & Protocol & \href{https://github.com/tlaplus/tlapm}{\tlaps{} Specification Repository} & 208 & 3 & (Safety + Refinement) \\
        \href{https://github.com/tlaplus/tlapm/blob/main/examples_draft/consensus/Voting.tla}{Consensus: Voting} & Protocol & \href{https://github.com/tlaplus/tlapm}{\tlaps{} Specification Repository} & 128 & 6 & (Safety + Refinement) \\
        \href{https://github.com/hengxin/tlaps-examples/blob/master/Record/Record.tla}{Record} & Protocol & \href{https://github.com/tlaplus/tlapm}{\tlaps{} Specification Repository} & 49 & 1 & (Refinement) \\
        \href{https://github.com/tlaplus/Examples/blob/master/specifications/Paxos/Consensus.tla}{Paxos: Consensus} & Protocol & \href{https://github.com/tlaplus/Examples}{\tlaplus{} Specification Repository} & 30 & 2 & (Safety + Liveness) \\
        \href{https://github.com/tlaplus/Examples/blob/master/specifications/Paxos/Voting.tla}{Paxos: Voting} & Protocol & \href{https://github.com/tlaplus/Examples}{\tlaplus{} Specification Repository} & 83 & 2 & (Safety + Refinement) \\
        \href{https://github.com/tlaplus/Examples/blob/master/specifications/PaxosHowToWinATuringAward/Consensus.tla}{Paxos lecture: Consensus} & Protocol & \href{https://github.com/tlaplus/Examples}{\tlaplus{} Specification Repository} & 27 & 1 & (Safety) \\
        \href{https://github.com/tlaplus/Examples/blob/master/specifications/PaxosHowToWinATuringAward/Voting.tla}{Paxos lecture: Voting} & Protocol & \href{https://github.com/tlaplus/Examples}{\tlaplus{} Specification Repository} & 101 & 2 & (Safety + Refinement) \\
        \href{https://github.com/tlaplus/Examples/blob/master/specifications/PaxosHowToWinATuringAward/Voting_proof.tla}{Paxos lecture: Voting proof} & Protocol & \href{https://github.com/tlaplus/Examples}{\tlaplus{} Specification Repository} & 78 & 3 & (Safety) \\
        \href{https://github.com/Starydark/Tencent-Paxos-TLA/blob/master/TPaxosWithProof.tla}{Tencent Paxos} & System & \href{https://github.com/tlaplus/Examples}{\tlaplus{} Specification Repository} & 189 & 2 & (Safety) \\
        \href{https://github.com/tlaplus/Examples/blob/master/specifications/byzpaxos/BPConProof.tla}{Byzantine Paxos: BPConProof} & Protocol & \href{https://github.com/tlaplus/Examples}{\tlaplus{} Specification Repository} & 400 & 3 & (Safety + Refinement) \\
        \href{https://github.com/tlaplus/Examples/blob/master/specifications/byzpaxos/Consensus.tla}{Byzantine Paxos: Consensus} & Protocol & \href{https://github.com/tlaplus/Examples}{\tlaplus{} Specification Repository} & 36 & 3 & (Safety + Liveness) \\
        \href{https://github.com/tlaplus/Examples/blob/master/specifications/byzpaxos/PConProof.tla}{Byzantine Paxos: PConProof} & Protocol & \href{https://github.com/tlaplus/Examples}{\tlaplus{} Specification Repository} & 94 & 1 & (Safety) \\
        \href{https://github.com/tlaplus/Examples/blob/master/specifications/byzpaxos/VoteProof.tla}{Byzantine Paxos: VoteProof} & Protocol & \href{https://github.com/tlaplus/Examples}{\tlaplus{} Specification Repository} & 147 & 4 & (Safety + Liveness + Refinement) \\
        \href{https://github.com/tlaplus/Examples/blob/master/specifications/bcastByz/bcastByz.tla}{Byzantine reliable broadcast} & Protocol & \href{https://github.com/tlaplus/Examples}{\tlaplus{} Specification Repository} & 130 & 5 & (Safety) \\
        \href{https://github.com/tlaplus/Examples/blob/master/specifications/bosco/bosco.tla}{Bosco Byzantine consensus} & Protocol & \href{https://github.com/tlaplus/Examples}{\tlaplus{} Specification Repository} & 140 & 5 & (Safety) \\
        \href{https://github.com/tlaplus/Examples/blob/master/specifications/dag-consensus/Sailfish.tla}{Sailfish BFT consensus} & Protocol & \href{https://github.com/tlaplus/Examples}{\tlaplus{} Specification Repository} & 200 & 3 & (Safety) \\
        \href{https://github.com/tlaplus/Examples/blob/master/specifications/NanoBlockchain/Nano.tla}{Nano blockchain} & System & \href{https://github.com/tlaplus/Examples}{\tlaplus{} Specification Repository} & 390 & 1 & (Safety) \\
        \href{https://github.com/kenmcmil/ivy/blob/master/examples/liveness/hybrid_reliable_broadcast_cisa.ivy}{Hybrid reliable broadcast} & Protocol & \href{https://github.com/kenmcmil/ivy}{Ivy examples} & 224 & 3 & (Safety + Liveness) \\
        \href{https://github.com/tlaplus/Examples/blob/master/specifications/TwoPhase/TwoPhase.tla}{Two-phase handshake} & Protocol & \href{https://github.com/tlaplus/Examples}{\tlaplus{} Specification Repository} & 51 & 1 & (Refinement) \\
        \href{https://github.com/tlaplus/Examples/blob/master/specifications/TwoPhase/TwoPhase_proof.tla}{Two-phase handshake (proof)} & Protocol & \href{https://github.com/tlaplus/Examples}{\tlaplus{} Specification Repository} & 50 & 1 & (Refinement) \\
        \href{https://github.com/tlaplus/Examples/blob/master/specifications/transaction_commit/PaxosCommit_proof.tla}{Paxos commit} & Protocol & \href{https://github.com/tlaplus/Examples}{\tlaplus{} Specification Repository} & 185 & 1 & (Safety) \\
        \href{https://github.com/tlaplus/Examples/blob/master/specifications/transaction_commit/TCommit_proof.tla}{Transaction commit} & Protocol & \href{https://github.com/tlaplus/Examples}{\tlaplus{} Specification Repository} & 46 & 1 & (Safety) \\
        \href{https://github.com/tlaplus/Examples/blob/master/specifications/transaction_commit/TwoPhase_proof.tla}{Two-phase commit} & Protocol & \href{https://github.com/tlaplus/Examples}{\tlaplus{} Specification Repository} & 112 & 2 & (Safety) \\
        \href{https://github.com/Vanlightly/s3-wal-collection/blob/06d5cc60990cf8e5995dc56f28e5d56128eca227/walgit/Walgit.tla}{Walgit WAL} & System & \href{https://github.com/Vanlightly/s3-wal-collection}{s3-wal-collection} & 558 & 7 & (Safety) \\
        \href{https://github.com/Vanlightly/s3-wal-collection/blob/06d5cc60990cf8e5995dc56f28e5d56128eca227/slatedb/SlateDBWAL.tla}{SlateDB WAL} & System & \href{https://github.com/Vanlightly/s3-wal-collection}{s3-wal-collection} & 546 & 4 & (Safety) \\
        \href{https://github.com/Vanlightly/s3-wal-collection/blob/06d5cc60990cf8e5995dc56f28e5d56128eca227/oswald/Oswald.tla}{OSWALD WAL} & System & \href{https://github.com/Vanlightly/s3-wal-collection}{s3-wal-collection} & 260 & 3 & (Safety) \\
        \href{https://github.com/tlaplus/Examples/blob/master/specifications/braf/BufferedRandomAccessFile.tla}{BufferedRandomAccessFile} & System & \href{https://github.com/tlaplus/Examples}{\tlaplus{} Specification Repository} & 334 & 14 & (Safety + Refinement) \\
        \href{https://github.com/tlaplus/tlapm/blob/main/examples/EWD840.tla}{EWD840 termination detection} & Protocol & \href{https://github.com/tlaplus/tlapm}{\tlaps{} Specification Repository} & 73 & 1 & (Safety) \\
        \href{https://github.com/tlaplus/Examples/blob/master/specifications/ewd840/EWD840_proof.tla}{EWD840} & Protocol & \href{https://github.com/tlaplus/Examples}{\tlaplus{} Specification Repository} & 121 & 2 & (Safety + Refinement) \\
        \href{https://github.com/tlaplus/Examples/blob/master/specifications/ewd840/SyncTerminationDetection_proof.tla}{EWD840: synchronous detection} & Protocol & \href{https://github.com/tlaplus/Examples}{\tlaplus{} Specification Repository} & 57 & 3 & (Safety + Liveness) \\
        \href{https://github.com/tlaplus/Examples/blob/91c22ea537853196ed1e03e9ad91693ec37642de/specifications/ewd687a/EWD687a_proof.tla}{Dijkstra--Scholten (EWD687a)} & Protocol & \href{https://github.com/tlaplus/Examples}{\tlaplus{} Specification Repository} & 132 & 3 & (Safety) \\
        \href{https://github.com/tlaplus/Examples/blob/master/specifications/ewd998/AsyncTerminationDetection_proof.tla}{EWD998: asynchronous detection} & Protocol & \href{https://github.com/tlaplus/Examples}{\tlaplus{} Specification Repository} & 71 & 3 & (Safety + Liveness) \\
        \href{https://github.com/tlaplus/Examples/blob/master/specifications/ewd998/EWD998PCal_proof.tla}{EWD998 (PlusCal)} & Protocol & \href{https://github.com/tlaplus/Examples}{\tlaplus{} Specification Repository} & 172 & 2 & (Safety + Refinement) \\
        \href{https://github.com/tlaplus/Examples/blob/master/specifications/ewd998/EWD998_proof.tla}{EWD998} & Protocol & \href{https://github.com/tlaplus/Examples}{\tlaplus{} Specification Repository} & 191 & 2 & (Safety + Refinement) \\
        \href{https://github.com/nano-o/Distributed-termination-detection/blob/main/Termination_proof.tla}{Termination detection} & Protocol & \href{https://github.com/tlaplus/Examples}{\tlaplus{} Specification Repository} & 69 & 1 & (Safety) \\
        \href{https://github.com/tlaplus/Examples/blob/master/specifications/MisraReachability/ParReachProofs.tla}{Misra reachability (parallel)} & Protocol & \href{https://github.com/tlaplus/Examples}{\tlaplus{} Specification Repository} & 143 & 1 & (Refinement) \\
        \href{https://github.com/tlaplus/Examples/blob/master/specifications/MisraReachability/ReachableProofs.tla}{Misra reachability (sequential)} & Protocol & \href{https://github.com/tlaplus/Examples}{\tlaplus{} Specification Repository} & 76 & 1 & (Safety) \\
        \href{https://github.com/tlaplus/Examples/blob/master/specifications/SpanningTree/SpanTree_proof.tla}{Spanning tree (SpanTree)} & Protocol & \href{https://github.com/tlaplus/Examples}{\tlaplus{} Specification Repository} & 48 & 1 & (Safety) \\
        \href{https://github.com/tlaplus/Examples/blob/master/specifications/spanning/spanning_proof.tla}{Spanning tree (spanning)} & Protocol & \href{https://github.com/tlaplus/Examples}{\tlaplus{} Specification Repository} & 44 & 1 & (Safety) \\
        \href{https://github.com/tlaplus/Examples/blob/master/specifications/tcp/tcp_proof.tla}{TCP} & Protocol & \href{https://github.com/tlaplus/Examples}{\tlaplus{} Specification Repository} & 264 & 3 & (Safety) \\
        \href{https://github.com/kenmcmil/ivy/blob/master/examples/liveness/alternating_bit_protocol.ivy}{Alternating-bit protocol} & Protocol & \href{https://github.com/kenmcmil/ivy}{Ivy examples} & 132 & 2 & (Safety + Liveness) \\
        \href{https://github.com/tlaplus/Examples/blob/master/specifications/FiniteMonotonic/CRDT_proof.tla}{FiniteMonotonic CRDT} & Protocol & \href{https://github.com/tlaplus/Examples}{\tlaplus{} Specification Repository} & 66 & 3 & (Safety + Liveness) \\
        \href{https://github.com/tlaplus/Examples/blob/master/specifications/MultiCarElevator/Elevator_proof.tla}{Multi-car elevator} & Protocol & \href{https://github.com/tlaplus/Examples}{\tlaplus{} Specification Repository} & 226 & 2 & (Safety) \\
        \href{https://github.com/tlaplus/Examples/blob/master/specifications/byihive/VoucherLifeCycle_proof.tla}{Voucher lifecycle} & System & \href{https://github.com/tlaplus/Examples}{\tlaplus{} Specification Repository} & 66 & 1 & (Safety) \\
        \href{https://github.com/tlaplus/Examples/blob/master/specifications/glowingRaccoon/clean_proof.tla}{PCR workflows: clean} & Protocol & \href{https://github.com/tlaplus/Examples}{\tlaplus{} Specification Repository} & 68 & 2 & (Safety) \\
        \href{https://github.com/tlaplus/Examples/blob/master/specifications/glowingRaccoon/stages_proof.tla}{PCR workflows: stages} & Protocol & \href{https://github.com/tlaplus/Examples}{\tlaplus{} Specification Repository} & 106 & 1 & (Safety) \\
        \midrule
        \textbf{Total} & & & & \multicolumn{2}{l@{}}{\textbf{\numRetiredTargets{}}} \\
\end{longtable}
\endgroup

\clearpage
\section{Detailed Benchmark Results}
\label{app:full-dataset-results}

Table~\ref{tab:per-target-results} shows the proof results for each target and model configuration.
Tables~\ref{tab:detailed-flash}--\ref{tab:detailed-wildfire} report \code{pass@3} success rates and total resource usage across all three runs.

\begingroup
\small
\setlength{\tabcolsep}{4pt}
\setlength{\LTleft}{0pt}
\setlength{\LTright}{0pt}
\setlength{\LTcapwidth}{\textwidth}
\begin{longtable}{@{\extracolsep{\fill}}lc*{12}{c}@{}}
    \caption{Accepted targets of the current problem set for each model and reasoning effort. S, L and R denote safety, liveness and refinement proofs, respectively.}
    \label{tab:per-target-results} \\
        \toprule
        \textbf{Target} & \textbf{Kind} & \multicolumn{3}{c}{\textbf{GPT-6.1 Sol}} & \multicolumn{3}{c}{\textbf{GPT-6 Astra}} & \multicolumn{3}{c}{\textbf{GPT-5.6 Terra}} & \multicolumn{3}{c}{\textbf{GPT-6 Luna}} \\
        \cmidrule(lr){3-5}\cmidrule(lr){6-8}\cmidrule(lr){9-11}\cmidrule(lr){12-14}
        & & max & high & low & max & high & low & max & high & low & max & high & low \\
        \midrule
    \endfirsthead
        \multicolumn{14}{@{}l}{\textit{Table~\thetable{} (continued)}} \\
        \toprule
        \textbf{Target} & \textbf{Kind} & \multicolumn{3}{c}{\textbf{GPT-6.1 Sol}} & \multicolumn{3}{c}{\textbf{GPT-6 Astra}} & \multicolumn{3}{c}{\textbf{GPT-5.6 Terra}} & \multicolumn{3}{c}{\textbf{GPT-6 Luna}} \\
        \cmidrule(lr){3-5}\cmidrule(lr){6-8}\cmidrule(lr){9-11}\cmidrule(lr){12-14}
        & & max & high & low & max & high & low & max & high & low & max & high & low \\
        \midrule
    \endhead
        \bottomrule
    \endfoot
        \multicolumn{14}{@{}l}{\textit{FLASH cache coherence protocol}} \\
        \code{TypeOK} & S & \checkmark & \checkmark & \checkmark & \checkmark & \checkmark & \checkmark & \checkmark &  &  & \checkmark &  &  \\
        \code{ReqProgress} & L & \checkmark & \checkmark & \checkmark & \checkmark & \checkmark & \checkmark &  &  &  &  &  &  \\
        \code{DirProgress} & L & \checkmark & \checkmark & \checkmark & \checkmark & \checkmark & \checkmark &  &  &  &  &  &  \\
        \code{UniProgress} & L & \checkmark & \checkmark & \checkmark & \checkmark & \checkmark & \checkmark &  &  &  &  &  &  \\
        \code{InvProgress} & L & \checkmark & \checkmark & \checkmark & \checkmark & \checkmark & \checkmark & \checkmark &  &  &  &  &  \\
        \code{RpProgress} & L & \checkmark & \checkmark & \checkmark & \checkmark & \checkmark & \checkmark & \checkmark &  &  &  &  &  \\
        \code{WbProgress} & L & \checkmark & \checkmark & \checkmark & \checkmark & \checkmark & \checkmark & \checkmark &  &  & \checkmark &  &  \\
        \code{ShWbProgress} & L & \checkmark & \checkmark & \checkmark & \checkmark & \checkmark & \checkmark & \checkmark &  &  & \checkmark &  &  \\
        \code{NakcProgress} & L & \checkmark & \checkmark & \checkmark & \checkmark & \checkmark & \checkmark & \checkmark &  &  & \checkmark &  &  \\
        \code{CacheDataProp} & S & \checkmark & \checkmark & \checkmark & \checkmark & \checkmark & \checkmark &  &  &  &  &  &  \\
        \code{MemDataProp} & S & \checkmark & \checkmark & \checkmark & \checkmark & \checkmark & \checkmark & \checkmark &  &  &  &  &  \\
        \code{ExCopyProp} & S & \checkmark & \checkmark & \checkmark & \checkmark & \checkmark & \checkmark & \checkmark &  &  &  &  &  \\
        \code{FwdGetPending} & S & \checkmark & \checkmark & \checkmark & \checkmark & \checkmark & \checkmark & \checkmark &  &  &  &  &  \\
        \code{FwdGetXPending} & S & \checkmark & \checkmark & \checkmark & \checkmark & \checkmark & \checkmark & \checkmark &  &  &  &  &  \\
        \code{InvalidationAckProp} & S & \checkmark & \checkmark & \checkmark & \checkmark & \checkmark & \checkmark & \checkmark &  &  &  &  &  \\
        \addlinespace
        \multicolumn{14}{@{}l}{\textit{ZooKeeper protocol (Zab)}} \\
        \code{Leadership1} & S &  &  &  & \checkmark &  &  &  &  &  &  &  &  \\
        \code{Leadership2} & S &  &  &  &  &  &  &  &  &  &  &  &  \\
        \code{PrefixConsistency} & S &  &  &  &  &  &  &  &  &  &  &  &  \\
        \code{Integrity} & S &  &  &  &  &  &  &  &  &  &  &  &  \\
        \code{Agreement} & S &  &  &  &  &  &  &  &  &  &  &  &  \\
        \code{TotalOrder} & S &  &  &  &  &  &  &  &  &  &  &  &  \\
        \code{LocalPrimaryOrder} & S &  &  &  &  &  &  &  &  &  &  &  &  \\
        \code{GlobalPrimaryOrder} & S &  &  &  &  &  &  &  &  &  &  &  &  \\
        \code{PrimaryIntegrity} & S &  &  &  &  &  &  &  &  &  &  &  &  \\
        \addlinespace
        \multicolumn{14}{@{}l}{\textit{Cahill's serializable snapshot isolation protocol}} \\
        \code{CahillSerializable} & S &  &  &  &  &  &  &  &  &  &  &  &  \\
        \addlinespace
        \multicolumn{14}{@{}l}{\textit{Ivy TLB}} \\
        \code{NoError} & S & \checkmark & \checkmark & \checkmark & \checkmark & \checkmark & \checkmark & \checkmark &  &  & \checkmark &  &  \\
        \code{NonStarvation} & L &  &  &  &  &  &  &  &  &  &  &  &  \\
        \addlinespace
        \multicolumn{14}{@{}l}{\textit{OpenAddressing}} \\
        \code{CompleteAsSafety} & S & \checkmark & \checkmark & \checkmark & \checkmark & \checkmark & \checkmark & \checkmark &  &  & \checkmark &  &  \\
        \code{Consistent} & S & \checkmark &  &  &  &  &  &  &  &  &  &  &  \\
        \code{Contains} & S & \checkmark &  &  &  &  &  &  &  &  &  &  &  \\
        \code{Duplicates} & S & \checkmark &  &  &  &  &  &  &  &  &  &  &  \\
        \code{Sorted} & S & \checkmark & \checkmark & \checkmark & \checkmark & \checkmark & \checkmark & \checkmark &  &  & \checkmark &  &  \\
        \addlinespace
        \multicolumn{14}{@{}l}{\textit{etcd Raft}} \\
        \code{LogInv} & S &  &  &  &  &  &  &  &  &  &  &  &  \\
        \code{CommittedIsDurableInv} & S &  &  &  &  &  &  &  &  &  &  &  &  \\
        \code{LeaderTermMaxIndexInv} & S & \checkmark &  &  &  &  &  &  &  &  &  &  &  \\
        \code{LeaderCompletenessInv} & S &  &  &  &  &  &  &  &  &  &  &  &  \\
        \code{LogMatchingInv} & S & \checkmark &  &  &  &  &  &  &  &  &  &  &  \\
        \code{LeaderUniqueness} & S & \checkmark & \checkmark & \checkmark & \checkmark & \checkmark & \checkmark &  &  &  &  &  &  \\
        \code{MoreUpToDateCorrectInv} & S &  &  &  &  &  &  &  &  &  &  &  &  \\
        \code{QuorumLogInv} & S &  &  &  &  &  &  &  &  &  &  &  &  \\
        \addlinespace
        \multicolumn{14}{@{}l}{\textit{HashiCorp Raft}} \\
        \code{LeaderCompleteness} & S &  &  &  &  &  &  &  &  &  &  &  &  \\
        \code{StateMachineSafety} & S &  &  &  &  &  &  &  &  &  &  &  &  \\
        \code{CommittedEntriesPreserved} & S &  &  &  &  &  &  &  &  &  &  &  &  \\
        \code{LogMatching} & S &  &  &  &  &  &  &  &  &  &  &  &  \\
        \code{ElectionSafety} & S &  &  &  &  &  &  &  &  &  &  &  &  \\
        \code{ConfigurationSafety} & S & \checkmark & \checkmark & \checkmark & \checkmark & \checkmark & \checkmark & \checkmark &  &  &  &  &  \\
        \addlinespace
        \multicolumn{14}{@{}l}{\textit{ZooKeeper implementation}} \\
        \code{Leadership1} & S &  &  &  & \checkmark &  &  &  &  &  &  &  &  \\
        \code{Leadership2} & S &  &  &  & \checkmark &  &  &  &  &  &  &  &  \\
        \code{PrefixConsistency} & S &  &  &  &  &  &  &  &  &  &  &  &  \\
        \code{Integrity} & S &  &  &  &  &  &  &  &  &  &  &  &  \\
        \code{Agreement} & S &  &  &  &  &  &  &  &  &  &  &  &  \\
        \code{TotalOrder} & S &  &  &  &  &  &  &  &  &  &  &  &  \\
        \code{LocalPrimaryOrder} & S &  &  &  &  &  &  &  &  &  &  &  &  \\
        \code{GlobalPrimaryOrder} & S &  &  &  &  &  &  &  &  &  &  &  &  \\
        \code{PrimaryIntegrity} & S &  &  &  &  &  &  &  &  &  &  &  &  \\
        \addlinespace
        \multicolumn{14}{@{}l}{\textit{MongoDB distributed transactions}} \\
        \code{SnapshotIsolation} & S &  &  &  &  &  &  &  &  &  &  &  &  \\
        \addlinespace
        \multicolumn{14}{@{}l}{\textit{CCF}} \\*
        \code{CommittedLogsAgree} & S &  &  &  &  &  &  &  &  &  &  &  &  \\
        \code{CommittedLogsAppendOnly} & S & \checkmark &  &  & \checkmark &  &  &  &  &  &  &  &  \\
        \addlinespace
        \multicolumn{14}{@{}l}{\textit{PirateShip}} \\*
        \code{TypeOK} & S & \checkmark &  & \checkmark & \checkmark &  &  &  &  &  &  &  &  \\
        \code{IndexBoundsInv} & S &  &  &  &  &  &  &  &  &  &  &  &  \\
        \code{BranchInv} & S &  &  &  &  &  &  &  &  &  &  &  &  \\
        \code{AuditBranchInv} & S &  &  &  &  &  &  &  &  &  &  &  &  \\
        \code{OneLeaderPerViewInv} & S & \checkmark & \checkmark & \checkmark & \checkmark & \checkmark & \checkmark &  &  &  &  &  &  \\
        \code{WellFormedBranchInv} & S &  &  &  & \checkmark &  &  &  &  &  &  &  &  \\
        \code{ViewMonotonicInv} & S & \checkmark & \checkmark & \checkmark & \checkmark & \checkmark & \checkmark &  &  &  &  &  &  \\
        \code{ViewStabilizationInv} & S & \checkmark & \checkmark & \checkmark & \checkmark & \checkmark & \checkmark &  &  &  &  &  &  \\
        \code{CommittedBranchAppendOnlyProp} & S &  &  &  &  &  &  &  &  &  &  &  &  \\
        \code{AuditedBranchAppendOnlyProp} & S &  &  &  &  &  &  &  &  &  &  &  &  \\
        \code{MonotonicAuditedIndexProp} & S & \checkmark & \checkmark & \checkmark & \checkmark & \checkmark & \checkmark &  &  &  &  &  &  \\
        \addlinespace
        \multicolumn{14}{@{}l}{\textit{Wildfire}} \\*
        \code{Refinement} & R &  &  &  &  &  &  &  &  &  &  &  &  \\
        \midrule
        \textbf{Accepted targets} & & \passThreeSolMaxAccepted & \passThreeSolHighAccepted & \passThreeSolLowAccepted & \passThreeAstraMaxAccepted & \passThreeAstraHighAccepted & \passThreeAstraLowAccepted & \passThreeTerraMaxAccepted & \passThreeTerraHighAccepted & \passThreeTerraLowAccepted & \passThreeLunaMaxAccepted & \passThreeLunaHighAccepted & \passThreeLunaLowAccepted \\
\end{longtable}
\endgroup

\begin{table*}[!htbp]
    \centering
    \small
    \caption{Results for the FLASH cache coherence protocol.}
    \label{tab:detailed-flash}
    \begin{tabular*}{\textwidth}{@{\extracolsep{\fill}}lcrrrrr@{}}
        \toprule
        \multirow{2}{*}{\textbf{Model}} & \multirow{2}{*}{\textbf{Effort}} & \multirow{2}{*}{\shortstack{\textbf{Success Rate}\\(pass@3)}} & \multicolumn{2}{c}{\textbf{Total Tokens (M)}} & \textbf{Total Cost} & \textbf{Total Agent Time} \\
        \cmidrule(lr){4-5}
        & & & Input & Output & (USD) & (min) \\
        \midrule
        \multirow{3}{*}{GPT-6.1 Sol} & max & \passThreeDetailFlashSolMax \\
         & high & \passThreeDetailFlashSolHigh \\
         & low & \passThreeDetailFlashSolLow \\
        \midrule
        \multirow{3}{*}{GPT-6 Astra} & max & \passThreeDetailFlashAstraMax \\
         & high & \passThreeDetailFlashAstraHigh \\
         & low & \passThreeDetailFlashAstraLow \\
        \midrule
        \multirow{3}{*}{GPT-5.6 Terra} & max & \passThreeDetailFlashTerraMax \\
         & high & \passThreeDetailFlashTerraHigh \\
         & low & \passThreeDetailFlashTerraLow \\
        \midrule
        \multirow{3}{*}{GPT-6 Luna} & max & \passThreeDetailFlashLunaMax \\
         & high & \passThreeDetailFlashLunaHigh \\
         & low & \passThreeDetailFlashLunaLow \\
        \bottomrule
    \end{tabular*}
\end{table*}

\begin{table*}[!htbp]
    \centering
    \small
    \caption{Results for the ZooKeeper protocol (Zab).}
    \label{tab:detailed-zab}
    \begin{tabular*}{\textwidth}{@{\extracolsep{\fill}}lcrrrrr@{}}
        \toprule
        \multirow{2}{*}{\textbf{Model}} & \multirow{2}{*}{\textbf{Effort}} & \multirow{2}{*}{\shortstack{\textbf{Success Rate}\\(pass@3)}} & \multicolumn{2}{c}{\textbf{Total Tokens (M)}} & \textbf{Total Cost} & \textbf{Total Agent Time} \\
        \cmidrule(lr){4-5}
        & & & Input & Output & (USD) & (min) \\
        \midrule
        \multirow{3}{*}{GPT-6.1 Sol} & max & \passThreeDetailZabSolMax \\
         & high & \passThreeDetailZabSolHigh \\
         & low & \passThreeDetailZabSolLow \\
        \midrule
        \multirow{3}{*}{GPT-6 Astra} & max & \passThreeDetailZabAstraMax \\
         & high & \passThreeDetailZabAstraHigh \\
         & low & \passThreeDetailZabAstraLow \\
        \midrule
        \multirow{3}{*}{GPT-5.6 Terra} & max & \passThreeDetailZabTerraMax \\
         & high & \passThreeDetailZabTerraHigh \\
         & low & \passThreeDetailZabTerraLow \\
        \midrule
        \multirow{3}{*}{GPT-6 Luna} & max & \passThreeDetailZabLunaMax \\
         & high & \passThreeDetailZabLunaHigh \\
         & low & \passThreeDetailZabLunaLow \\
        \bottomrule
    \end{tabular*}
\end{table*}

\begin{table*}[!htbp]
    \centering
    \small
    \caption{Results for Cahill's serializable snapshot isolation protocol.}
    \label{tab:detailed-cahill}
    \begin{tabular*}{\textwidth}{@{\extracolsep{\fill}}lcrrrrr@{}}
        \toprule
        \multirow{2}{*}{\textbf{Model}} & \multirow{2}{*}{\textbf{Effort}} & \multirow{2}{*}{\shortstack{\textbf{Success Rate}\\(pass@3)}} & \multicolumn{2}{c}{\textbf{Total Tokens (M)}} & \textbf{Total Cost} & \textbf{Total Agent Time} \\
        \cmidrule(lr){4-5}
        & & & Input & Output & (USD) & (min) \\
        \midrule
        \multirow{3}{*}{GPT-6.1 Sol} & max & \passThreeDetailCahillSolMax \\
         & high & \passThreeDetailCahillSolHigh \\
         & low & \passThreeDetailCahillSolLow \\
        \midrule
        \multirow{3}{*}{GPT-6 Astra} & max & \passThreeDetailCahillAstraMax \\
         & high & \passThreeDetailCahillAstraHigh \\
         & low & \passThreeDetailCahillAstraLow \\
        \midrule
        \multirow{3}{*}{GPT-5.6 Terra} & max & \passThreeDetailCahillTerraMax \\
         & high & \passThreeDetailCahillTerraHigh \\
         & low & \passThreeDetailCahillTerraLow \\
        \midrule
        \multirow{3}{*}{GPT-6 Luna} & max & \passThreeDetailCahillLunaMax \\
         & high & \passThreeDetailCahillLunaHigh \\
         & low & \passThreeDetailCahillLunaLow \\
        \bottomrule
    \end{tabular*}
\end{table*}

\begin{table*}[!htbp]
    \centering
    \small
    \caption{Results for Ivy TLB.}
    \label{tab:detailed-ivy-tlb}
    \begin{tabular*}{\textwidth}{@{\extracolsep{\fill}}lcrrrrr@{}}
        \toprule
        \multirow{2}{*}{\textbf{Model}} & \multirow{2}{*}{\textbf{Effort}} & \multirow{2}{*}{\shortstack{\textbf{Success Rate}\\(pass@3)}} & \multicolumn{2}{c}{\textbf{Total Tokens (M)}} & \textbf{Total Cost} & \textbf{Total Agent Time} \\
        \cmidrule(lr){4-5}
        & & & Input & Output & (USD) & (min) \\
        \midrule
        \multirow{3}{*}{GPT-6.1 Sol} & max & \passThreeDetailIvyTlbSolMax \\
         & high & \passThreeDetailIvyTlbSolHigh \\
         & low & \passThreeDetailIvyTlbSolLow \\
        \midrule
        \multirow{3}{*}{GPT-6 Astra} & max & \passThreeDetailIvyTlbAstraMax \\
         & high & \passThreeDetailIvyTlbAstraHigh \\
         & low & \passThreeDetailIvyTlbAstraLow \\
        \midrule
        \multirow{3}{*}{GPT-5.6 Terra} & max & \passThreeDetailIvyTlbTerraMax \\
         & high & \passThreeDetailIvyTlbTerraHigh \\
         & low & \passThreeDetailIvyTlbTerraLow \\
        \midrule
        \multirow{3}{*}{GPT-6 Luna} & max & \passThreeDetailIvyTlbLunaMax \\
         & high & \passThreeDetailIvyTlbLunaHigh \\
         & low & \passThreeDetailIvyTlbLunaLow \\
        \bottomrule
    \end{tabular*}
\end{table*}

\begin{table*}[!htbp]
    \centering
    \small
    \caption{Results for OpenAddressing.}
    \label{tab:detailed-open-addressing}
    \begin{tabular*}{\textwidth}{@{\extracolsep{\fill}}lcrrrrr@{}}
        \toprule
        \multirow{2}{*}{\textbf{Model}} & \multirow{2}{*}{\textbf{Effort}} & \multirow{2}{*}{\shortstack{\textbf{Success Rate}\\(pass@3)}} & \multicolumn{2}{c}{\textbf{Total Tokens (M)}} & \textbf{Total Cost} & \textbf{Total Agent Time} \\
        \cmidrule(lr){4-5}
        & & & Input & Output & (USD) & (min) \\
        \midrule
        \multirow{3}{*}{GPT-6.1 Sol} & max & \passThreeDetailOpenAddressingSolMax \\
         & high & \passThreeDetailOpenAddressingSolHigh \\
         & low & \passThreeDetailOpenAddressingSolLow \\
        \midrule
        \multirow{3}{*}{GPT-6 Astra} & max & \passThreeDetailOpenAddressingAstraMax \\
         & high & \passThreeDetailOpenAddressingAstraHigh \\
         & low & \passThreeDetailOpenAddressingAstraLow \\
        \midrule
        \multirow{3}{*}{GPT-5.6 Terra} & max & \passThreeDetailOpenAddressingTerraMax \\
         & high & \passThreeDetailOpenAddressingTerraHigh \\
         & low & \passThreeDetailOpenAddressingTerraLow \\
        \midrule
        \multirow{3}{*}{GPT-6 Luna} & max & \passThreeDetailOpenAddressingLunaMax \\
         & high & \passThreeDetailOpenAddressingLunaHigh \\
         & low & \passThreeDetailOpenAddressingLunaLow \\
        \bottomrule
    \end{tabular*}
\end{table*}

\begin{table*}[!htbp]
    \centering
    \small
    \caption{Results for etcd Raft.}
    \label{tab:detailed-etcd}
    \begin{tabular*}{\textwidth}{@{\extracolsep{\fill}}lcrrrrr@{}}
        \toprule
        \multirow{2}{*}{\textbf{Model}} & \multirow{2}{*}{\textbf{Effort}} & \multirow{2}{*}{\shortstack{\textbf{Success Rate}\\(pass@3)}} & \multicolumn{2}{c}{\textbf{Total Tokens (M)}} & \textbf{Total Cost} & \textbf{Total Agent Time} \\
        \cmidrule(lr){4-5}
        & & & Input & Output & (USD) & (min) \\
        \midrule
        \multirow{3}{*}{GPT-6.1 Sol} & max & \passThreeDetailEtcdSolMax \\
         & high & \passThreeDetailEtcdSolHigh \\
         & low & \passThreeDetailEtcdSolLow \\
        \midrule
        \multirow{3}{*}{GPT-6 Astra} & max & \passThreeDetailEtcdAstraMax \\
         & high & \passThreeDetailEtcdAstraHigh \\
         & low & \passThreeDetailEtcdAstraLow \\
        \midrule
        \multirow{3}{*}{GPT-5.6 Terra} & max & \passThreeDetailEtcdTerraMax \\
         & high & \passThreeDetailEtcdTerraHigh \\
         & low & \passThreeDetailEtcdTerraLow \\
        \midrule
        \multirow{3}{*}{GPT-6 Luna} & max & \passThreeDetailEtcdLunaMax \\
         & high & \passThreeDetailEtcdLunaHigh \\
         & low & \passThreeDetailEtcdLunaLow \\
        \bottomrule
    \end{tabular*}
\end{table*}

\begin{table*}[!htbp]
    \centering
    \small
    \caption{Results for HashiCorp Raft.}
    \label{tab:detailed-hashicorp}
    \begin{tabular*}{\textwidth}{@{\extracolsep{\fill}}lcrrrrr@{}}
        \toprule
        \multirow{2}{*}{\textbf{Model}} & \multirow{2}{*}{\textbf{Effort}} & \multirow{2}{*}{\shortstack{\textbf{Success Rate}\\(pass@3)}} & \multicolumn{2}{c}{\textbf{Total Tokens (M)}} & \textbf{Total Cost} & \textbf{Total Agent Time} \\
        \cmidrule(lr){4-5}
        & & & Input & Output & (USD) & (min) \\
        \midrule
        \multirow{3}{*}{GPT-6.1 Sol} & max & \passThreeDetailHashicorpSolMax \\
         & high & \passThreeDetailHashicorpSolHigh \\
         & low & \passThreeDetailHashicorpSolLow \\
        \midrule
        \multirow{3}{*}{GPT-6 Astra} & max & \passThreeDetailHashicorpAstraMax \\
         & high & \passThreeDetailHashicorpAstraHigh \\
         & low & \passThreeDetailHashicorpAstraLow \\
        \midrule
        \multirow{3}{*}{GPT-5.6 Terra} & max & \passThreeDetailHashicorpTerraMax \\
         & high & \passThreeDetailHashicorpTerraHigh \\
         & low & \passThreeDetailHashicorpTerraLow \\
        \midrule
        \multirow{3}{*}{GPT-6 Luna} & max & \passThreeDetailHashicorpLunaMax \\
         & high & \passThreeDetailHashicorpLunaHigh \\
         & low & \passThreeDetailHashicorpLunaLow \\
        \bottomrule
    \end{tabular*}
\end{table*}

\begin{table*}[!htbp]
    \centering
    \small
    \caption{Results for the ZooKeeper implementation.}
    \label{tab:detailed-zookeeper}
    \begin{tabular*}{\textwidth}{@{\extracolsep{\fill}}lcrrrrr@{}}
        \toprule
        \multirow{2}{*}{\textbf{Model}} & \multirow{2}{*}{\textbf{Effort}} & \multirow{2}{*}{\shortstack{\textbf{Success Rate}\\(pass@3)}} & \multicolumn{2}{c}{\textbf{Total Tokens (M)}} & \textbf{Total Cost} & \textbf{Total Agent Time} \\
        \cmidrule(lr){4-5}
        & & & Input & Output & (USD) & (min) \\
        \midrule
        \multirow{3}{*}{GPT-6.1 Sol} & max & \passThreeDetailZooKeeperSolMax \\
         & high & \passThreeDetailZooKeeperSolHigh \\
         & low & \passThreeDetailZooKeeperSolLow \\
        \midrule
        \multirow{3}{*}{GPT-6 Astra} & max & \passThreeDetailZooKeeperAstraMax \\
         & high & \passThreeDetailZooKeeperAstraHigh \\
         & low & \passThreeDetailZooKeeperAstraLow \\
        \midrule
        \multirow{3}{*}{GPT-5.6 Terra} & max & \passThreeDetailZooKeeperTerraMax \\
         & high & \passThreeDetailZooKeeperTerraHigh \\
         & low & \passThreeDetailZooKeeperTerraLow \\
        \midrule
        \multirow{3}{*}{GPT-6 Luna} & max & \passThreeDetailZooKeeperLunaMax \\
         & high & \passThreeDetailZooKeeperLunaHigh \\
         & low & \passThreeDetailZooKeeperLunaLow \\
        \bottomrule
    \end{tabular*}
\end{table*}

\begin{table*}[!htbp]
    \centering
    \small
    \caption{Results for MongoDB distributed transactions.}
    \label{tab:detailed-mongodb}
    \begin{tabular*}{\textwidth}{@{\extracolsep{\fill}}lcrrrrr@{}}
        \toprule
        \multirow{2}{*}{\textbf{Model}} & \multirow{2}{*}{\textbf{Effort}} & \multirow{2}{*}{\shortstack{\textbf{Success Rate}\\(pass@3)}} & \multicolumn{2}{c}{\textbf{Total Tokens (M)}} & \textbf{Total Cost} & \textbf{Total Agent Time} \\
        \cmidrule(lr){4-5}
        & & & Input & Output & (USD) & (min) \\
        \midrule
        \multirow{3}{*}{GPT-6.1 Sol} & max & \passThreeDetailMongoSolMax \\
         & high & \passThreeDetailMongoSolHigh \\
         & low & \passThreeDetailMongoSolLow \\
        \midrule
        \multirow{3}{*}{GPT-6 Astra} & max & \passThreeDetailMongoAstraMax \\
         & high & \passThreeDetailMongoAstraHigh \\
         & low & \passThreeDetailMongoAstraLow \\
        \midrule
        \multirow{3}{*}{GPT-5.6 Terra} & max & \passThreeDetailMongoTerraMax \\
         & high & \passThreeDetailMongoTerraHigh \\
         & low & \passThreeDetailMongoTerraLow \\
        \midrule
        \multirow{3}{*}{GPT-6 Luna} & max & \passThreeDetailMongoLunaMax \\
         & high & \passThreeDetailMongoLunaHigh \\
         & low & \passThreeDetailMongoLunaLow \\
        \bottomrule
    \end{tabular*}
\end{table*}

\begin{table*}[!htbp]
    \centering
    \small
    \caption{Results for CCF.}
    \label{tab:detailed-ccf}
    \begin{tabular*}{\textwidth}{@{\extracolsep{\fill}}lcrrrrr@{}}
        \toprule
        \multirow{2}{*}{\textbf{Model}} & \multirow{2}{*}{\textbf{Effort}} & \multirow{2}{*}{\shortstack{\textbf{Success Rate}\\(pass@3)}} & \multicolumn{2}{c}{\textbf{Total Tokens (M)}} & \textbf{Total Cost} & \textbf{Total Agent Time} \\
        \cmidrule(lr){4-5}
        & & & Input & Output & (USD) & (min) \\
        \midrule
        \multirow{3}{*}{GPT-6.1 Sol} & max & \passThreeDetailCcfSolMax \\
         & high & \passThreeDetailCcfSolHigh \\
         & low & \passThreeDetailCcfSolLow \\
        \midrule
        \multirow{3}{*}{GPT-6 Astra} & max & \passThreeDetailCcfAstraMax \\
         & high & \passThreeDetailCcfAstraHigh \\
         & low & \passThreeDetailCcfAstraLow \\
        \midrule
        \multirow{3}{*}{GPT-5.6 Terra} & max & \passThreeDetailCcfTerraMax \\
         & high & \passThreeDetailCcfTerraHigh \\
         & low & \passThreeDetailCcfTerraLow \\
        \midrule
        \multirow{3}{*}{GPT-6 Luna} & max & \passThreeDetailCcfLunaMax \\
         & high & \passThreeDetailCcfLunaHigh \\
         & low & \passThreeDetailCcfLunaLow \\
        \bottomrule
    \end{tabular*}
\end{table*}

\begin{table*}[!htbp]
    \centering
    \small
    \caption{Results for PirateShip.}
    \label{tab:detailed-pirateship}
    \begin{tabular*}{\textwidth}{@{\extracolsep{\fill}}lcrrrrr@{}}
        \toprule
        \multirow{2}{*}{\textbf{Model}} & \multirow{2}{*}{\textbf{Effort}} & \multirow{2}{*}{\shortstack{\textbf{Success Rate}\\(pass@3)}} & \multicolumn{2}{c}{\textbf{Total Tokens (M)}} & \textbf{Total Cost} & \textbf{Total Agent Time} \\
        \cmidrule(lr){4-5}
        & & & Input & Output & (USD) & (min) \\
        \midrule
        \multirow{3}{*}{GPT-6.1 Sol} & max & \passThreeDetailPirateShipSolMax \\
         & high & \passThreeDetailPirateShipSolHigh \\
         & low & \passThreeDetailPirateShipSolLow \\
        \midrule
        \multirow{3}{*}{GPT-6 Astra} & max & \passThreeDetailPirateShipAstraMax \\
         & high & \passThreeDetailPirateShipAstraHigh \\
         & low & \passThreeDetailPirateShipAstraLow \\
        \midrule
        \multirow{3}{*}{GPT-5.6 Terra} & max & \passThreeDetailPirateShipTerraMax \\
         & high & \passThreeDetailPirateShipTerraHigh \\
         & low & \passThreeDetailPirateShipTerraLow \\
        \midrule
        \multirow{3}{*}{GPT-6 Luna} & max & \passThreeDetailPirateShipLunaMax \\
         & high & \passThreeDetailPirateShipLunaHigh \\
         & low & \passThreeDetailPirateShipLunaLow \\
        \bottomrule
    \end{tabular*}
\end{table*}

\begin{table*}[!htbp]
    \centering
    \small
    \caption{Results for Wildfire.}
    \label{tab:detailed-wildfire}
    \begin{tabular*}{\textwidth}{@{\extracolsep{\fill}}lcrrrrr@{}}
        \toprule
        \multirow{2}{*}{\textbf{Model}} & \multirow{2}{*}{\textbf{Effort}} & \multirow{2}{*}{\shortstack{\textbf{Success Rate}\\(pass@3)}} & \multicolumn{2}{c}{\textbf{Total Tokens (M)}} & \textbf{Total Cost} & \textbf{Total Agent Time} \\
        \cmidrule(lr){4-5}
        & & & Input & Output & (USD) & (min) \\
        \midrule
        \multirow{3}{*}{GPT-6.1 Sol} & max & \passThreeDetailWildfireSolMax \\
         & high & \passThreeDetailWildfireSolHigh \\
         & low & \passThreeDetailWildfireSolLow \\
        \midrule
        \multirow{3}{*}{GPT-6 Astra} & max & \passThreeDetailWildfireAstraMax \\
         & high & \passThreeDetailWildfireAstraHigh \\
         & low & \passThreeDetailWildfireAstraLow \\
        \midrule
        \multirow{3}{*}{GPT-5.6 Terra} & max & \passThreeDetailWildfireTerraMax \\
         & high & \passThreeDetailWildfireTerraHigh \\
         & low & \passThreeDetailWildfireTerraLow \\
        \midrule
        \multirow{3}{*}{GPT-6 Luna} & max & \passThreeDetailWildfireLunaMax \\
         & high & \passThreeDetailWildfireLunaHigh \\
         & low & \passThreeDetailWildfireLunaLow \\
        \bottomrule
    \end{tabular*}
\end{table*}

\clearpage

\lstdefinelanguage{tlasnippet}{
  morekeywords={ASSUME,PROVE,THEOREM,LEMMA,PROOF,BY,DEF,ONLY,OBVIOUS,QED,SUFFICES,CASE,PICK,NEW,LET,IN,IF,THEN,ELSE,EXCEPT,UNCHANGED,ENABLED,UNION,SUBSET,DOMAIN,CONSTANT,INSTANCE,LOCAL},
  sensitive=true,
  morecomment=[l]{\\*},
}
\renewcommand{\lstlistingname}{Example}
\lstdefinestyle{tlasnippet}{
  language=tlasnippet,
  basicstyle=\ttfamily\small,
  keywordstyle=\bfseries,
  commentstyle=\color{black!60}\itshape,
  columns=fullflexible,
  keepspaces=true,
  breaklines=true,
  breakatwhitespace=true,
  breakindent=0pt,
  showstringspaces=false,
  upquote=true,
  literate={"}{{\char34\relax}}1 {~}{{$\sim$}}1,
  frame=lines,
  framesep=4pt,
  captionpos=b,
  abovecaptionskip=0pt,
  aboveskip=10pt,
  belowskip=10pt,
}

\lstnewenvironment{tlaexample}[1][]
  {\lstset{style=tlasnippet,#1}}
  {}

\section{Accepted Proofs, Incomplete Proofs, and Disproved Targets}
\label{app:proof-examples}

This appendix shows simplified versions of proofs submitted by agents in the benchmark runs of Section~\ref{sec:evaluation}, including both accepted proofs and incomplete proofs. We also include one disproved target found during task validation (Section~\ref{sec:problem-set-validation}). In \tlaps{}, a \code{BY} clause lists facts and proof hints to justify a proof step. A hint can choose a backend prover or pick a proof strategy: \code{SMTT} calls an SMT solver, \code{Isa} calls Isabelle, \code{PTL} calls the temporal logic prover, \code{ExpandENABLED} hints the backend to expand the \code{ENABLED} operator. We drop some auxiliary steps to keep the examples short.

\subsection{Accepted Proofs}
\label{app:accepted-proofs}

\para{FLASH: eventual message delivery.} FLASH has eight liveness targets concerning request completion and message delivery. Example~\ref{lst:flash-liveness} shows a proof of \code{WbProgressCorrect}, which states that a pending writeback message is eventually delivered under the assumption of weak fairness: if the action remains enabled, it eventually takes a step~\citep{lamport1994tla}. The action \code{NI\_Wb} delivers the message, and the specification \code{FairSpec} assumes weak fairness of \code{NI\_Wb}. Step \code{<1>1} shows that \code{NI\_Wb} can be taken while the message is pending, using the type invariant \code{TypeOK}. Step \code{<1>2} shows that if \code{NI\_Wb} is taken, the message will be delivered. Step \code{<1>3} extracts the fairness assumption from \code{FairSpec}, and step \code{<1>4} shows that \code{TypeOK} always holds by reusing the proved target \code{TypeCorrect}. The other seven liveness targets follow the same pattern or chain several such steps through helper lemmas.

\begin{tlaexample}[caption={Accepted proof of the FLASH liveness target by GPT-6.1 Sol at high effort (simplified).},label={lst:flash-liveness}]
\* (target and action, from the read-only task files)
WbProgress == (WbMsg.Cmd = "WB_Wb") ~> (WbMsg.Cmd = "WB_None")
NI_Wb ==
    /\ WbMsg.Cmd = "WB_Wb"
    /\ WbMsg' = [Cmd |-> "WB_None", Proc |-> Undefined, Data |-> Undefined]
    /\ ...  \* (omitted: directory and memory updates)

THEOREM WbProgressCorrect == FairSpec => WbProgress
<1>1. TypeOK /\ (WbMsg.Cmd # "WB_None") => ENABLED <<NI_Wb>>_vars
  BY SMTT("r30"), ExpandENABLED
  DEF NI_Wb, TypeOK, vars, pendVars, fwdVars, RP_CMD, WB_CMD, SHWB_CMD, NAKC_CMD
<1>2. <<NI_Wb>>_vars => (WbMsg.Cmd = "WB_None")'
  BY SMTT("r10") DEF NI_Wb, vars
<1>3. FairSpec => WF_vars(NI_Wb)
  BY Isa DEF FairSpec, Fairness
<1>4. FairSpec => []TypeOK
  BY TypeCorrect, SMTT("r10") DEF FairSpec, Spec
<1>5. QED BY <1>1, <1>2, <1>3, <1>4, PTL DEF WbProgress
\end{tlaexample}

\para{etcd Raft: leader uniqueness.} The target \code{LeaderUniqueness} states that two leaders in the same term are the same server. Example~\ref{lst:etcd-accepted} shows the accepted proof by GPT-6 Astra at max effort, and Figure~\ref{fig:task-example} shows a simplified version of this task. The auxiliary invariant \code{AgElectionInv} treats each granted vote as a ballot \code{<<voter, term, candidate>>}, wherever the vote is currently stored: in a pending or delivered vote response, or in the \code{votesGranted} set of a candidate or leader. \code{AgAgreement} states that a voter's ballots in one term name at most one candidate. \code{AgPendingBound} and \code{AgDurableBound} tie each ballot to the voter's own term and vote. They use durable state for delivered and collected votes, so the invariant still holds after a server crashes and restarts. Given the invariant, the target follows from quorum intersection: two leaders in the same term each hold a quorum of votes, a voter in both quorums has a ballot for each leader, and \code{AgAgreement} makes the two leaders equal. The agent proves that every action preserves the invariant with one lemma per action.

\begin{tlaexample}[caption={Accepted proof of the etcd Raft target \code{LeaderUniqueness} by GPT-6 Astra at max effort (simplified).},label={lst:etcd-accepted}]
\* Vote certificates are retained in pending messages, delivered messages,
\* and the vote sets of active candidates and leaders.
AgVoteMsgs(bag) ==
  {m \in DOMAIN bag : m.mtype = RequestVoteResponse /\ m.mvoteGranted}
AgMsgBallots(bag) == {<<m.msource, m.mterm, m.mdest>> : m \in AgVoteMsgs(bag)}
AgPending == AgMsgBallots(pendingMessages)
AgSent == AgMsgBallots(messages)
AgGranted ==
  UNION {{<<v, currentTerm[i], i>> : v \in votesGranted[i]} : i \in {j \in Server : AgActive(j)}}
AgBallots == AgPending \cup AgSent \cup AgGranted
\* Pending votes are protected by volatile state.  Delivered and collected
\* votes are protected by durable state, so their protection survives a crash.
AgPendingBound == \A b \in AgPending :
  /\ b[2] <= currentTerm[b[1]]
  /\ b[2] = currentTerm[b[1]] => votedFor[b[1]] = b[3]
AgDurableBound == \A b \in AgSent \cup AgGranted :
  /\ b[2] <= durableState[b[1]].currentTerm
  /\ b[2] = durableState[b[1]].currentTerm
     => durableState[b[1]].votedFor = b[3]
\* A voter has at most one candidate among certificates for the same term.
AgAgreement == \A b, c \in AgBallots :
  b[1] = c[1] /\ b[2] = c[2] => b[3] = c[3]
AgElectionInv ==
  /\ AgBasic /\ ...  \* (AgBasic includes state[i] = Leader =>
                     \*  votesGranted[i] \in Quorum(InitServer))
  /\ AgPendingBound /\ AgDurableBound /\ AgAgreement

LEMMA AgElectionSafetyConsequence == AgElectionInv => LeaderUniqueness
<1>1. ASSUME AgElectionInv, NEW i \in Server, NEW j \in Server,
             currentTerm[i] = currentTerm[j], state[i] = Leader, state[j] = Leader
      PROVE i = j
  <2>1. votesGranted[i] \in Quorum(InitServer)
        /\ votesGranted[j] \in Quorum(InitServer)
    BY <1>1, SMTT("r5") DEF AgElectionInv, AgBasic
  <2>2. PICK v \in InitServer :
          v \in votesGranted[i] /\ v \in votesGranted[j]
    BY <2>1, AgQuorumIntersection, FiniteServers
  <2>3. <<v, currentTerm[i], i>> \in AgBallots
        /\ <<v, currentTerm[j], j>> \in AgBallots
    BY <1>1, <2>2, SMTT("r5") DEF AgBallots, AgGranted, AgActive
  <2>4. QED BY <1>1, <2>3, SMTT("r5") DEF AgElectionInv, AgAgreement
<1>2. QED BY <1>1 DEF LeaderUniqueness

\* (omitted: AgElectionInit and AgElectionStep, the initial-state and step lemmas)
LEMMA AgElectionAlways == Spec => []AgElectionInv
BY AgElectionInit, AgElectionStep, PTL DEF Spec
THEOREM LeaderUniquenessTheorem == Spec => []LeaderUniqueness
BY ONLY AgElectionAlways, AgElectionSafetyConsequence, PTL
\end{tlaexample}

\subsection{Proof Failures}
\label{app:incomplete-proofs}
\label{app:detailed-failure-analysis}

An accepted proof of a safety target \code{Spec => []P} (Example~\ref{lst:etcd-accepted}) usually requires discovering and proving auxiliary invariants and helper lemmas, and deriving \code{P} from these proof facts. Liveness targets (Example~\ref{lst:flash-liveness}) additionally require progress lemmas connecting fairness assumptions with intermediate state machine stages. Any missing segment in the proof chain will fail the task.

The two most common failure categories in Table~\ref{tab:failure-categories} are undecomposed proof attempts and incomplete proofs. In undecomposed attempts, agents submit entire targets directly to backends. Agents may do so before developing a proof strategy or after their own proof attempts stall. We illustrate this with an Ivy TLB example below. Incomplete proofs reveal difficulties in discovering useful invariants, establishing proposed invariants or lemmas, and connecting established results to the target property. We present one example for each of these three cases below.

\para{Undecomposed liveness proof: Ivy TLB.} GPT-6 Astra at max effort proved the TLB safety target with an inductive invariant \code{AGInv}, and all 30 of its helper lemmas passed verification. For the liveness target \code{NonStarvation}, however, the agent submitted a single call to the temporal-logic backend \code{PTL} (Example~\ref{lst:tlb-undecomposed}), which must derive the target from the definitions of \code{Spec} and \code{NonStarvation} alone. A liveness proof instead needs progress lemmas showing that, under the fairness assumptions in \code{Spec}, the processor's loops terminate and the locks it waits for are eventually released. The agent did attempt such a proof in scratch modules alongside the safety proof, and checked its finiteness and control-flow steps. However, it could not prove that the actions in the fairness assumptions are enabled, which is required to apply fairness. It kept these attempts out of the submission and ended the session with the one-step proof it had written at the start, which the backend rejects in under a second.

\begin{tlaexample}[caption={Undecomposed proof of the Ivy TLB liveness target \code{NonStarvation} by GPT-6 Astra at max effort (simplified). \code{Safety} is accepted; the one-step proof of \code{Liveness} fails verification.},label={lst:tlb-undecomposed}]
\* (target and fairness assumptions, from the read-only task files)
Spec ==
  /\ SafetySpec  \* (Init /\ [][Next]_vars)
  /\ \A p \in Processor : WF_vars(Step(p))
  /\ \A p \in Processor : SF_vars(BootStep(p))
  /\ \A p \in Processor : SF_vars(AcquirePmapLock(p))
  /\ \A p \in Processor : SF_vars(AcquireResponderActionLock(p))
ProcessorMakesProgress(p) == pc[p] \in {MainCheck, ResponderClearActionNeeded}
NonStarvation == \A p \in Processor : TRUE ~> ProcessorMakesProgress(p)

\* (helpers for Safety, all verified; proofs and the per-action lemmas omitted)
AGInv == AGType /\ AGPmap /\ AGAction /\ AGControl /\ AGCoverage
         /\ AGCoherent /\ ~error
LEMMA AGInitial == Init => AGInv
LEMMA AGInduction == AGInv /\ [Next]_vars => AGInv'
LEMMA AGInvariant == SafetySpec => []AGInv

THEOREM Safety == SafetySpec => []NoError
BY AGInvariant, PTL DEF AGInv, NoError

THEOREM Liveness == Spec => NonStarvation
BY PTL DEF Spec, NonStarvation
\end{tlaexample}

\para{Missing invariant for one case: OpenAddressing.} In the OpenAddressing model, writers insert fingerprints into a hash table. A writer probes up to \code{L} slots, starting from the fingerprint's home slot \code{idx(f, 0)}. If it finds neither the fingerprint nor a free slot, it triggers an eviction, which moves table entries into external sorted sequences. GPT-6 Astra at max effort proves two of the five targets and verifies all 50 of its helper lemmas, but leaves \code{Consistent} (Example~\ref{lst:openaddressing-incomplete}) and other proofs incomplete. The proof splits into two cases. In the first case, \code{OAHashScale > 0} and \code{L > 0}, so distinct fingerprints have distinct home slots, every insertion succeeds at its first probe, and no eviction occurs. The invariant \code{OADirectInv} describes this behavior, and this case is verified. The second case covers hash collisions and evictions, where probing and eviction actually take place. The agent's comment notes that this case needs a stronger invariant, but the step cites only three existing invariants, and the SMT solver fails to prove it. The proof therefore covers only configurations without collisions or evictions.

\begin{tlaexample}[caption={Incomplete proof of the OpenAddressing target \code{Consistent} by GPT-6 Astra at max effort (simplified). Step \code{<1>2} fails verification.},label={lst:openaddressing-incomplete}]
\* (when OAHashScale > 0, distinct fingerprints have distinct home slots idx(f, 0))
OAHashScale == (K-1) \div (max(fps)-min(fps))

OADirectInv ==
 /\ ...  \* (omitted: types and control states)
 /\ external = <<>>
 /\ newexternal = <<>>
 /\ evict = FALSE
 /\ index = [w \in Writer |-> 0]
 /\ \A i \in 1..K : table[i] # empty => i = idx(table[i],0)
 /\ ...

THEOREM Spec => []Consistent
<1>1. OAHashScale > 0 /\ L > 0 => (Spec => []Consistent)
  <2>1. OADirectInv => Consistent BY OADirectSafety
  <2>2. QED BY <2>1, OADirectInvariant, PTL
<1>2. OAHashScale = 0 \/ L = 0 => (Spec => []Consistent)
  \* The collision case also needs preservation of the table and scan invariant.
  BY OAControlInvariant, OATypesInvariant, OASortedInvariant, SMTT("r1")
<1>3. OAHashScale \in Nat /\ L \in Nat
  BY OAHashScaleType, OAAssumption, SMTT("r10")
<1>4. QED BY <1>1, <1>2, <1>3, SMTT("r10")
\end{tlaexample}

\para{Missing progress lemmas: Ivy TLB.} GPT-6 Astra at high effort proved the TLB safety target with an inductive invariant and verified all 29 of its helper lemmas. Its proof of the liveness target \code{NonStarvation} (Example~\ref{lst:tlb-incomplete}) combined this invariant and the specification's fairness assumptions and tried to prove the liveness goal. However, the agent did not state the progress lemmas needed to show that the locks are eventually released and the processor's loops terminate. As a result, the final step failed verification, and a comment the agent left in the proof listed these missing steps.

\begin{tlaexample}[caption={Incomplete proof of the Ivy TLB liveness target \code{NonStarvation} by GPT-6 Astra at high effort (simplified). Step \code{<2>} \code{QED} fails verification, as do steps \code{<2>1}--\code{<2>3}.},label={lst:tlb-incomplete}]
\* (target and fairness assumptions, from the read-only task files)
Spec ==
  /\ SafetySpec  \* (Init /\ [][Next]_vars)
  /\ \A p \in Processor : WF_vars(Step(p))
  /\ \A p \in Processor : SF_vars(BootStep(p))
  /\ \A p \in Processor : SF_vars(AcquirePmapLock(p))
  /\ \A p \in Processor : SF_vars(AcquireResponderActionLock(p))
ProcessorMakesProgress(p) == pc[p] \in {MainCheck, ResponderClearActionNeeded}
NonStarvation == \A p \in Processor : TRUE ~> ProcessorMakesProgress(p)

\* (helpers, all verified; proofs and most conjuncts omitted)
AgentInv == AgentTypes /\ AgentActivity /\ AgentPmapLock /\ ... /\ NoError
LEMMA AgentInitial == Init => AgentInv
LEMMA AgentInductive == AgentInv /\ [Next]_vars => AgentInv'

THEOREM Liveness == Spec => NonStarvation
<1>1. SafetySpec => []AgentInv
  BY AgentInitial, AgentInductive, PTL DEF SafetySpec
<1>2. ASSUME NEW p \in Processor
      PROVE Spec => (TRUE ~> ProcessorMakesProgress(p))
  <2>1. Spec => []AgentInv
    BY <1>1 DEF Spec
  <2>2. Spec => WF_vars(Step(p))
    BY <1>2 DEF Spec
  <2>3. Spec => /\ SF_vars(BootStep(p))
                /\ SF_vars(AcquirePmapLock(p))
                /\ SF_vars(AcquireResponderActionLock(p))
    BY <1>2 DEF Spec
  \* Remaining obligation: termination of the two todo loops and
  \* eventual release of the locks on which this processor can wait.
  <2> QED BY <2>1, <2>2, <2>3, PTL DEF ProcessorMakesProgress
<1> QED BY <1>2 DEF NonStarvation
\end{tlaexample}

\para{Unproved invariant: Cahill SSI.} GPT-6 Astra at max effort stated a candidate invariant \code{CSInvariant} for SSI and proved that it implies serializability and holds at \code{Init} (step \code{<1>1} of Example~\ref{lst:cahill-unproved}), but it did not prove the inductive step \code{<1>2}, which shows that every action preserves the invariant. Instead, the agent tested the invariant with TLC by simulating executions from \code{Init}. Passing these tests only shows that \code{CSInvariant} holds in the sampled reachable states of a small model, while the proof additionally requires it to be inductive.

\begin{tlaexample}[caption={Incomplete proof of the Cahill SSI target by GPT-6 Astra at max effort (simplified), with the TLC model it used to test the invariant. Step \code{<1>2} fails verification. The TLC model starts from \code{Init}, so it checks the invariant only on reachable states.},label={lst:cahill-unproved}]
CSInvariant ==
  /\ CSStateTypes /\ CSCore /\ CSWaitSafe
  /\ CSHistoryConditions(history, inConflict, outConflict)
\* (helpers, both verified; proofs omitted)
LEMMA CSInvariantInitial == Init => CSInvariant
LEMMA CSInvariantSafety == CSInvariant => CahillSerializable(history)

THEOREM CahillSerializableCorrect == Spec => []CahillSerializable(history)
<1>1. Init => CSInvariant BY CSInvariantInitial, SMTT("r5")
<1>2. CSInvariant /\ Next => CSInvariant'
  BY SMTT("r5") DEF CSInvariant, Next
<1>3. CSInvariant /\ UNCHANGED allvars => CSInvariant'
  ...  \* (omitted: proof for stuttering steps)
<1>4. CSInvariant => CahillSerializable(history) BY CSInvariantSafety, SMTT("r5")
<1>5. QED BY <1>1, <1>2, <1>3, <1>4, PTL DEF Spec

\* (the agent's TLC model, scratch/MCCahill.tla and MCCahill.cfg)
MCInv == CSInvariant /\ CahillSerializable(history)
INIT Init
NEXT Next
INVARIANT MCInv
\end{tlaexample}

\subsection{Disproved Targets}
\label{app:disproved-targets}

\para{SlateDB WAL: a counterexample to the target.} A task can also end with a counterexample when the target is false (Section~\ref{sec:problem-def}). In a run before the evaluation of Section~\ref{sec:evaluation}, GPT-6 Astra at max effort proved one of the four targets of the original SlateDB WAL model and reported that \code{ConsistentMachineData} does not hold and submitted a 58-transition TLC counterexample. Example~\ref{lst:slatedb-counterexample} shows the agent's summary of the trace. We reproduced the trace, reported it upstream, and revised the model; Appendix~\ref{app:spec-fixes} describes the repair, after which Astra proves all four targets.

\begin{tlaexample}[caption={Counterexample summary submitted by GPT-6 Astra at max effort for the SlateDB WAL target \code{ConsistentMachineData}.},label={lst:slatedb-counterexample}]
The essential schedule is:

1. Writer w1 becomes ready with a fence at WAL ID 1.
2. Writer w2 chooses WAL ID 2 as its fence position. Before it claims its epoch,
   w1 appends v1 at ID 2 and v2 at ID 3.
3. w2 claims epoch 2 and passes the boundary check, but delays writing its fence.
4. w3 claims epoch 3, writes its fence at ID 4, validates successfully, and
   starts replaying. Pause it before replaying ID 2.
5. w4 claims epoch 4, writes its fence at ID 5, replays the full log, and commits
   snapshot 5 containing <<v1, v2>>. The latest manifest now has ID 6 and
   replayAfterWalId = 6.
6. The garbage collector deletes the data entry at ID 2, leaving ID 3 present.
7. The delayed w2 writes a fence at the now-vacant ID 2.
8. w3 skips the new fence at ID 2, reads v2 at ID 3, skips its own fence at
   ID 4, and reaches READY with machine data <<v2>>.

The final history is <<[walId |-> 2, value |-> v1], [walId |-> 3, value |-> v2]>>.
Since v1 # v2, PrefixOf(<<v2>>, auxWrittenEntries) is false.
Thus Spec => []ConsistentMachineData is false under the supplied context.
\end{tlaexample}

\clearpage

\section{Representative Specification Fixes}
\label{app:spec-fixes}

Section~\ref{sec:problem-def} summarizes the issues we found in source specifications during task construction. Table~\ref{tab:spec-fixes} lists representative issues. For each issue, we collect counterexamples and check each repair with model checking. We describe two of these cases in detail below.

\begin{table}[H]
    \centering
    \small
    \caption{Representative specification issues and their repairs.}
    \label{tab:spec-fixes}
    \begin{tabularx}{\textwidth}{@{}>{\raggedright\arraybackslash}p{0.15\textwidth}>{\raggedright\arraybackslash}X>{\raggedright\arraybackslash}p{0.22\textwidth}>{\raggedright\arraybackslash}p{0.12\textwidth}@{}}
        \toprule
        \textbf{Specification} & \textbf{Issue} & \textbf{Repair} & \textbf{Status} \\
        \midrule
        OpenAddressing & With a single fingerprint, the divisor \code{max(fps) - min(fps)} in the hash rescaling is zero. & Require a finite set of at least two fingerprints. & Fixed upstream \\
        MongoDB & \code{vars} omits five of the 25 state variables, so stuttering steps can change them. & Add them to \code{vars}. & Fixed upstream \\
        Walgit WAL & Overlapping \code{CASE} guards allow replay to skip a checkpoint and omit committed data. Reordering the branches exposes a violation hidden by TLC's default order. & Make replay and retry guards mutually exclusive. & Reported; no response \\
        MongoDB & One transaction can start at two routers with different read timestamps, violating \code{SnapshotIsolation} (Section~\ref{sec:problem-def}). & Check all routers before a transaction starts. & Fixed upstream \\
        Sailfish & Correct nodes accept a Byzantine leader's proposal without a predecessor reference or a no-vote certificate, violating \code{Agreement}. & Check the predecessor reference or no-vote certificate before admission. & Fixed upstream \\
        SlateDB WAL & A delayed writer's fence, written after garbage collection, is skipped during replay, and \code{ConsistentMachineData} fails. & Validate the writer's epoch after replay. & Reported; no response \\
        B-tree & \code{Next} has no action for splitting an inner node, so an insertion can deadlock. & Not repaired; the five safety targets do not concern insertion progress. & Acknowledged upstream \\
        HashiCorp Raft & Replication acknowledgments report the follower's whole log length instead of the replicated prefix. & Acknowledge the replicated prefix, as the implementation does. & Fixed upstream \\
        etcd Raft & \code{LeaderCompletenessInv} compares leaders with the term that created an entry instead of the term that committed it. & Compare committed prefixes with later elections. & Fixed upstream \\
        \bottomrule
    \end{tabularx}
\end{table}

\para{etcd Raft: a bug in the theorem.} Raft's leader completeness property says that an entry committed in some term appears in the log of every leader of a later term~\citep{ongaro2014search}. The original \code{LeaderCompletenessInv} (Example~\ref{lst:etcd-fix}) instead compares a leader's term with the term in which the entry was created. This theorem fails as Raft allows an entry to be committed late. The agent reported a counterexample with 47 transitions. We correct the theorem to talk about the committed term. We use \code{commitHistory} to record the term and committed log prefix whenever a server's commit index advances, and \code{electionHistory} to record each leader's term and log at election time.

\begin{tlaexample}[caption={Correction of the etcd Raft target \code{LeaderCompletenessInv}.},label={lst:etcd-fix}]
-LeaderCompletenessInv ==
-    \A i \in Server :
-        LET committed == Committed(i) IN
-        \A idx \in 1..Len(committed) :
-            LET entry == log[i][idx] IN
-            \A l \in CurrentLeaders :
-                currentTerm[l] > entry.term =>
-                log[l][idx] = entry
+LeaderCompletenessInv ==
+    \A c \in commitHistory, e \in electionHistory :
+        c.term < e.term => IsPrefix(c.entries, e.entries)
\end{tlaexample}

\para{SlateDB WAL: a bug in the model.} GPT-6 Astra disproved \code{ConsistentMachineData} on the original SlateDB WAL model with a 58-transition TLC counterexample (Example~\ref{lst:slatedb-counterexample}). After garbage collection deletes a DATA entry, a delayed writer writes a FENCE at the same position. Another writer replaying from an older snapshot skips the FENCE and enters READY without recovering the deleted data. We reported the trace upstream. Our revision adds a validation step (Example~\ref{lst:slatedb-fix}): after replay, the writer reloads the latest log metadata and stops if a newer writer has taken over. The target properties are unchanged. On the revised model, Astra proves all four targets.

\begin{tlaexample}[caption={Repair of the SlateDB WAL model (abridged diff, line breaks added).},label={lst:slatedb-fix}]
 ReplayWAL(w) ==
     ...
           CASE id = wNextWalId[w] ->
-                    /\ wState' = [wState EXCEPT ![w] = READY]
+                    /\ wState' = [wState EXCEPT ![w] = VALIDATE_AFTER_REPLAY]
                     /\ wReplayId' = [wReplayId EXCEPT ![w] = 0]
     ...
+ValidateEpochAfterReplay(w) ==
+    /\ wState[w] = VALIDATE_AFTER_REPLAY
+    /\ CASE DoRefresh(w) ->
+                /\ RefreshManifest(w)
+                /\ UNCHANGED wState
+         [] ~DoRefresh(w) /\ (wManifestLoad[w].valid = FALSE
+                              \/ wEpoch[w] < wManifest[w].writerEpoch) ->
+                /\ wState' = [wState EXCEPT ![w] = FENCED]
+                /\ ManifestRefreshed(w)
+         [] ...  \* (omitted: a newer local epoch leads to ILLEGAL_STATE)
+         [] ~DoRefresh(w) /\ wManifestLoad[w].valid
+                          /\ wEpoch[w] = wManifest[w].writerEpoch ->
+                /\ wState' = [wState EXCEPT ![w] = READY]
+                /\ ManifestRefreshed(w)
+    /\ UNCHANGED <<storeVars, gcVars, auxVars, wEpoch,
+                   wNextWalId, wReplayId, wMachineData>>
     ...
 Next ==
     ...
         \/ ReplayWAL(w)
+        \/ ValidateEpochAfterReplay(w)
\end{tlaexample}

\clearpage
\begingroup
\renewcommand*\ttdefault{txtt}
\section{Agent Prompt and Skills}
\label{app:prompts}

We reproduce the proof construction prompt and the two agent skills supplied with the \harness{}. The prompt is shown for \code{etcd\_raft.tla}.

\subsection{Proof Construction Prompt}
\label{app:proof-prompt}

The following prompt specifies the editable regions, available proof libraries, checker commands, and requirements for an accepted proof.

\begin{promptbox}[frametitle={Proof Construction Prompt}]
\begin{lstlisting}[style=benchmarkprompt]
# Task

Prove every target theorem in etcd_raft.tla. This is the only submitted file and represents one complete module task. Every supplied TLA+ context file in the workspace is read-only.

You may create temporary wrappers, configs, traces, and other scratch files for local TLC or Apalache exploration. They are not submitted. Grading reconstructs a clean workspace from etcd_raft.tla and the canonical read-only context, so the final task must not import or depend on scratch files.

Only edit inside the `\* BEGIN AGENT HELPERS` ... `\* END AGENT HELPERS` region and each identified AGENT PROOF region (`\* BEGIN AGENT PROOF <proof-unit-id>` ... `\* END AGENT PROOF <proof-unit-id>`). Do not change the proof-unit IDs on the marker lines. You may choose the proof strategy and reuse any earlier sibling theorem once it is proved.

Put fresh same-module operator definitions, module-level USE DEF or HIDE DEF directives, and fully proved named helper lemmas in AGENT HELPERS. Put each target theorem's proof only in its matching identified proof region.

# Proving and iterating

Audited official proof libraries are available at /opt/proof-libraries/tlapm and $COMMUNITY_LIB. Allowed modules are listed in $TLAPS_PROOF_LIBRARY_CATALOG. Inside AGENT HELPERS you may add LOCAL <alias> == INSTANCE <module> from that catalog, but you must not modify library files or use `WITH`.

You may run:

~~~sh
/opt/tlapm/bin/tlapm -I /opt/proof-libraries/tlapm -I "$COMMUNITY_LIB" etcd_raft.tla
~~~

Keep refining the module until this authoritative check reports PASS:

~~~sh
check_proof_bin etcd_raft.tla --mode proof-from-scratch
~~~

Use this command without overriding `--timeout` or `--check-cpus`; it inherits the run's frozen verification policy. If a check reports a configuration error, follow its `RETRY-COMMAND` and obtain a proof-check result before submitting. Raw TLAPM and `--sany-only` do not check the complete submission contract.

The checker reports a raw result for every target and every helper theorem used by a target. A raw PASS counts only when all local theorem dependencies are also trusted, so a proof cannot rely on an omitted or circular sibling.

Run the checker in the foreground and wait for that tool execution to finish. For a long-running tool call, use the tool's blocking wait facility rather than repeatedly launching checks or polling files. Each independent self-check has the same module budget as final grading; use `--check-session DIR` only when intentionally resuming one interrupted check of the unchanged candidate. A `NOT_STARTED` result means the module budget expired before that unit was checked, not that its proof failed.

For a quick syntax check, run:

~~~sh
check_proof_bin etcd_raft.tla --mode proof-from-scratch --sany-only
~~~

If you must stop early, leave your best partial proofs in the module and run the authoritative check on that final file. Fix reported integrity violations, including edits to fixed whitespace, and rerun it after any correction. Independently trusted proof units still receive credit when the submission satisfies the integrity checks.

# Constraints

- Do not change the module header, imports, marker lines, theorem statements, theorem order, or any text outside the editable regions.
- Do not modify or replace supplied dependency modules, and do not make the final task depend on scratch modules.
- Do not add EXTENDS, nested modules, CONSTANT, VARIABLE, ASSUME, ASSUMPTION, or AXIOM declarations. INSTANCE is allowed only in the official-library form described above.
- Every helper `LEMMA` or `THEOREM` must be named and fully proved.
- Never use PROOF OMITTED or bare OMITTED, and never create a circular helper.
- Bound the SMT/Z3 backend with a deterministic budget SMTT("rN"), capped at SMTT("r30"). Do not use the wall-clock forms SMT or SMTT(n).
- New operator and lemma names must be fresh. Do not shadow names from the task or its context.
- Do not browse or fetch external material. Work only from the files in this workspace and your knowledge of TLA+/TLAPS.
\end{lstlisting}
\end{promptbox}

\subsection{Agent Skills}
\label{app:skills}

Runs with the \harness{} also receive the two agent skills described in Section~\ref{sec:proof-tools}, reproduced below.

\begin{promptbox}[frametitle={Skill: tlaps-proof-hints}]
\begin{lstlisting}[style=benchmarkprompt]
---
name: tlaps-proof-hints
description: Resolve TLAPS proof failures involving theorem instances from modules with assumptions. Use when a citation such as BY I!Thm does not close a goal because prefixed and unprefixed imported operators are treated as different symbols.
---

# Use a theorem from an instantiated module

## Symptom

A citation such as `BY I!Thm, NA` does not close the goal even though `I!Thm` appears to prove it and `NA` appears to supply the corresponding local assumption.

## Cause

When a module with assumptions is instantiated as `I == INSTANCE M`, the instantiated assumptions become hypotheses of its theorem instances. If an assumption mentions an operator imported through `EXTENDS`, the instance re-exports that operator under the `I!` prefix. TLAPS treats the prefixed and unprefixed names as distinct internal symbols, so backends do not automatically relate `I!Pos(N)` to `Pos(N)`.

## Fix

Prove a bridge theorem equating the prefixed operator with its local counterpart and cite it alongside the instantiated theorem:

```tla
---- MODULE Lib ----
EXTENDS Integers
Pos(n) == n > 0
====


---- MODULE A ----
EXTENDS Lib
CONSTANT N
ASSUME NA == Pos(N)

THEOREM Thm == N > 0
  BY NA DEF Pos
====


---- MODULE B ----
EXTENDS Lib
CONSTANT N
ASSUME NA == Pos(N)

I == INSTANCE A

THEOREM Bridge == \A n : I!Pos(n) = Pos(n)
  BY DEF I!Pos, Pos

THEOREM Test == N > 0
\*  BY I!Thm, NA            \* FAILS
  BY I!Thm, NA, Bridge      \* WORKS
====
```

Source: [tlaplus/tlapm#279](https://github.com/tlaplus/tlapm/pull/279)
\end{lstlisting}
\end{promptbox}

\begin{promptbox}[frametitle={Skill: tla-inductive-invariant-validation}]
\begin{lstlisting}[style=benchmarkprompt]
---
name: tla-inductive-invariant-validation
description: Validate candidate inductive invariants for TLA+ safety proofs with Apalache or TLC. Use when designing or debugging an invariant, especially when TLAPS cannot discharge initiation, consecution, or safety-implication obligations.
---

# Validate a TLA+ inductive invariant

Use a model checker to falsify a candidate invariant before investing in its TLAPS proof. Treat a successful finite check as evidence only: prove the invariant in TLAPS, and finish with the benchmark's authoritative `check_proof_bin` command.

## Run the three Apalache checks

Shape the candidate as `IndInv == TypeOK /\ H`, where `TypeOK` contains one finite-domain membership clause for every variable and `H` contains the additional facts needed for induction. Given a target module `Target`, a typed wrapper `APTarget.tla`, and a safety predicate `Safety`, run:

```sh
apalache-mc check --config=APTarget.cfg --init=Init --inv=IndInv --length=0 APTarget.tla
apalache-mc check --config=APTarget.cfg --init=IndInv --inv=IndInv --length=1 APTarget.tla
apalache-mc check --config=APTarget.cfg --init=IndInv --inv=Safety --length=0 APTarget.tla
```

Interpret these checks as:

1. `Init => IndInv`: legitimate initial states satisfy the candidate.
2. `IndInv /\ Next => IndInv'`: one step preserves the candidate.
3. `IndInv => Safety`: the candidate is strong enough to prove the goal.

The checks starting from `--init=IndInv` require the candidate to assign every variable. A bare safety predicate can instead produce an assignment error such as `x' is used before it is assigned`. Write the missing `TypeOK` if the stripped target has none. Bound `Nat`-typed values to small finite sets, for example `num \in [P -> 0..4]`, and widen a bound before trusting a counterexample that may depend on it.

When a check fails, read the counterexample. Apalache writes `violation1.tla` in a timestamped directory under `_apalache-out/<module>.tla/` unless redirected with `--out-dir` or `--run-dir`; the `Check the trace in: ...` log line gives the exact path.

- A failure of the first check means the candidate excludes a legitimate initial state.
- A failure of the second check gives a state satisfying `IndInv` and a successor violating it. The trailing `InvariantViolation` formula identifies the broken conjunct and usually exposes the fact missing from `H`.
- A failure of the third check means the candidate does not imply the safety goal.

Strengthen or correct the candidate from the trace, then rerun all three checks.

## Build a non-invasive Apalache wrapper

Keep tool-specific annotations out of `Target.tla`. Re-declare its constants and variables with Apalache types in a separate wrapper, then instantiate the target:

```tla
---- MODULE APTarget ----
EXTENDS Naturals

CONSTANT
  \* @type: Int;
  N

VARIABLE
  \* @type: Set(Int);
  active

INSTANCE Target

TypeOK == active \in SUBSET (0..N)
H == TRUE
IndInv == TypeOK /\ H

====
```

Match every re-declared name exactly so implicit `INSTANCE` substitution binds it. Put each `\* @type:` annotation between the `CONSTANT` or `VARIABLE` keyword and the name it annotates, including every name originally declared in a multi-name block. A comment above the keyword is not accepted and leads to `Expected a type annotation for VARIABLE <name>`.

Define `IndInv` in the wrapper. Fix finite constants in `APTarget.cfg`, for example:

```tla
INIT Init
NEXT Next

CONSTANT N = 5
```

Alternatively, use `--cinit`. For an uninterpreted constant, define a finite value operator such as `JugVal` in the wrapper and substitute it with `Jug <- JugVal` in the configuration. If a polymorphic helper such as `Sum`, `Range`, or `ToSet` defeats type checking, shadow it in the wrapper with a monomorphic version.

Only if the wrapper cannot be made to typecheck, annotate a private scratch copy:

```sh
mkdir -p scratch
cp Target.tla scratch/Candidate.tla
```

Never make the submitted solution extend or instantiate the wrapper or scratch module. Do not alter the target or any dependency as part of model-checking exploration.

## Fall back to TLC

Use TLC when Apalache cannot type the target through a wrapper. Create a separate `MCTarget.tla` that extends `Target` and defines `TypeOK`, `H`, and `IndInv`, with `MCTarget.cfg` beside it. TLC is not on `PATH`; run:

```sh
java -cp /opt/sany/lib/tla2tools.jar tlc2.TLC -deadlock -config MCTarget.cfg MCTarget.tla
```

Check the inductive step by making the candidate its own initial predicate:

```tla
INIT IndInv
NEXT Next
INVARIANT IndInv
```

This checks `IndInv /\ Next => IndInv'` because `IndInv` is an ordinary invariant of the specification started in `IndInv`. Check `Init => IndInv` separately on the real specification with `SPECIFICATION Spec` and `INVARIANT IndInv`. Include `Safety` as a conjunct of `IndInv` to check that the candidate implies it.

Use the same enumerable `TypeOK /\ H` shape as for Apalache. TLC first builds the states satisfying `TypeOK` and then filters them with `H`, so start with the smallest useful model. To reduce enumeration:

- Move a conjunct of `H` immediately after the `TypeOK` clauses that bind the variables it mentions, allowing TLC to prune earlier.
- Sample large domains with `RandomSubset(k, S)` from the `Randomization` module. Rewrite `v \subseteq W` as `v \in SUBSET W` and sample it with `RandomSetOfSubsets(k, p, W)`.

Aim for a few dozen initial states. A TLC violation supplies the same useful two-state signal as Apalache. Once random sampling is involved, a run with no violation proves nothing; rerun it about a dozen times and abandon a run whose search depth grows beyond roughly a dozen steps.

See Leslie Lamport's [Using TLC to Check Inductive Invariance](https://lamport.azurewebsites.net/tla/inductive-invariant.pdf) for the TLC technique.

Never cite an Apalache or TLC verdict as justification for a TLAPS proof step.
\end{lstlisting}
\end{promptbox}
\endgroup

\end{document}